\documentclass[10pt]{article}

\usepackage{speakermemr1,times}
\usepackage[utf8]{inputenc}
\usepackage[T1]{fontenc}
\usepackage{microtype}
\usepackage{amsmath,amssymb,mathtools}
\usepackage{booktabs}
\usepackage{multirow}
\usepackage{array}
\usepackage{tabularx}
\usepackage{graphicx}
\usepackage{float}
\usepackage{placeins}
\usepackage{flafter}
\usepackage{tikz}
\usetikzlibrary{positioning,arrows.meta,fit,calc,shapes.geometric}
\usepackage{xcolor}
\usepackage{colortbl}
\usepackage{enumitem}
\usepackage{url}
\usepackage{xurl}
\usepackage{hyperref}
\usepackage{cleveref}
\usepackage{makecell}
\usepackage{pifont}
\usepackage{fvextra}
\usepackage[most]{tcolorbox}
\fvset{fontsize=\scriptsize,breaklines=true,breakanywhere=true,frame=single,framesep=1.5mm}

\hypersetup{
  colorlinks=true,
  linkcolor=black,
  citecolor=black,
  urlcolor=black
}

\newcommand{\method}{\textsc{SpeakerMem-R1}}

\newcommand{\best}[1]{\textbf{#1}}
\newcommand{\second}[1]{\underline{#1}}

\tcbset{
  promptbox/.style={
    enhanced jigsaw,
    breakable,
    listing only,
    colback=black!2.5!white,
    colframe=black!22,
    colbacktitle=black!8!white,
    coltitle=black!72,
    fonttitle=\sffamily\bfseries\footnotesize,
    attach boxed title to top left={xshift=3.0mm,yshift*=-\tcboxedtitleheight/2},
    boxed title style={
      colback=black!8!white,
      colframe=black!35,
      boxrule=.35pt,
      arc=1.1mm,
      outer arc=1.1mm,
      left=2.0mm,
      right=2.0mm,
      top=.75mm,
      bottom=.75mm
    },
    boxrule=.55pt,
    arc=2.0mm,
    outer arc=2.0mm,
    left=3.0mm,
    right=3.0mm,
    top=4.8mm,
    bottom=2.8mm,
    before skip=9pt,
    after skip=8pt,
    listing options={
      basicstyle=\ttfamily\fontsize{8.0}{10.0}\selectfont,
      breaklines=true,
      breakatwhitespace=false,
      literate=*%
        {[SYSTEM PROMPT]}{{{\color{black!72}\sffamily\bfseries [SYSTEM PROMPT]}}}{15}%
        {[USER TEMPLATE]}{{{\color{black!72}\sffamily\bfseries [USER TEMPLATE]}}}{15}%
        {[RAW EXCERPTS]}{{{\color{black!72}\sffamily\bfseries [RAW EXCERPTS]}}}{14}%
        {[MEMORY SLICE]}{{{\color{black!72}\sffamily\bfseries [MEMORY SLICE]}}}{14},
      columns=fullflexible,
      keepspaces=true,
      showstringspaces=false,
      upquote=true,
      tabsize=2,
      aboveskip=0pt,
      belowskip=0pt,
      xleftmargin=0pt,
      xrightmargin=0pt
    }
  }
}
\newcommand{\promptinput}[2]{%
  \tcbinputlisting{promptbox,title={#1},listing file={#2}}%
}

\smrfinalcopy
\renewcommand{\headrulewidth}{0pt}
\renewcommand{\footrulewidth}{0pt}

\microtypesetup{protrusion=false}
\title{SpeakerMem-R1: Speaker-Centered Dual-Track Memory for Multi-Party Dialogue}
\author{%
  \begin{tabular}{c}
    \normalfont Haobo Zheng, Tan Tang\textsuperscript{\textdagger}, Yan Chen, Weijie Wang, Yingcai Wu\\
    \normalfont State Key Lab of CAD\&CG, Zhejiang University
  \end{tabular}
}

\definecolor{arxivlightblue}{RGB}{239,247,255}
\renewenvironment{abstract}{%
  \begin{tcolorbox}[
    enhanced,
    breakable,
    width=\linewidth,
    colback=arxivlightblue,
    colframe=arxivlightblue,
    frame hidden,
    boxrule=0pt,
    arc=1.5mm,
    outer arc=1.5mm,
    left=4mm,
    right=4mm,
    top=3mm,
    bottom=3mm,
    drop shadow={black!18!white}
  ]%
  \begin{center}\textbf{\sffamily Abstract}\end{center}\vspace{-0.5em}
}{%
  \end{tcolorbox}\par\medskip
}

\begin{document}
\maketitle
\begingroup
\renewcommand{\thefootnote}{\fnsymbol{footnote}}
\footnotetext[2]{Corresponding author.}
\endgroup

\begin{abstract}
Long-term conversational memory in multi-party settings requires more than retrieving relevant content from long-term conversations: it must distinguish who said what, whom each statement concerns, how individuals perceive one another, what information is shared by the group, and how states change over time. Recent studies on multi-party dialogue benchmarks show that existing general-purpose LLM memory systems tend to lose person and group relations or struggle to integrate clues distributed across members, groups, and time. Together, these issues reveal two core bottlenecks: message attribution and relational understanding in multi-party dialogue, and state reconstruction from interleaved histories. To address both, we propose \textbf{\method{}}: its dual-track memory stores speaker-labeled verbatim messages and derived states organized into person-level and group-level views, then combines evidence from both tracks by entity, event, and time at query time. To reduce attribution and update errors during structured memory construction while enabling local deployment, we train Writer-R1 with SpeakerLevenshtein and speaker-conditioned GRPO. On GroupMemBench, SocialMemBench, and EverMemBench, \method{} achieves binary accuracies of 47.9\%, 69.2\%, and 61.9\%, respectively. The scores are 3.3, 12.4, and 9.4 percentage points over the best results of mainstream frameworks evaluated on each benchmark, respectively. On the publicly reported EverMemBench leaderboard from EverMind-AI, \method{} achieves 62.33\%, the best reported result among the latest state-of-the-art frameworks. It also achieves 70.85\% on all 1,986 LoCoMo questions, which we use as a two-person long-term conversation boundary test. In a controlled evaluation of 305 questions, RL raises the SFT Writer's mean accuracy from 57.38\% to 68.20\% under a frozen query/answer pipeline. We report both binary accuracy and token-F1, and ablations show that the verbatim and structured tracks, as well as person-level and group-level views, are complementary under the standardized evaluation interface.
\par\medskip
\noindent\textbf{Project Page:} \href{https://2022hpsk.github.io/SpeakerMemR1}{\textnormal{https://2022hpsk.github.io/SpeakerMemR1}}\\
\textbf{GitHub:} \href{https://github.com/2022hpsk/SpeakerMemR1}{\textnormal{https://github.com/2022hpsk/SpeakerMemR1}}
\end{abstract}

\section{Introduction}
\label{sec:introduction}

Long-term conversational memory enables language agents to retain facts, preferences, and social relations across sessions~\citep{zhong2023memorybank,park2023generative,packer2023memgpt,wu2025longmemeval}. Existing work mainly targets single-user or two-person histories, splitting, compressing, indexing, and retrieving conversations by relevance~\citep{maharana2024locomo,wu2025longmemeval}. Multi-party group chats additionally contain speaker relations, reply structure, cross-topic branches, and state revisions, so they cannot be flattened into a message stream that is simply compressed in order~\citep{ghosal2019dialoguegcn,li2020molweni}. GroupMemBench, SocialMemBench, and EverMemBench further show that general-purpose memory systems degrade substantially in multi-party or long-term conversation settings, while BM25 or dense retrieval can remain competitive in some configurations~\citep{yang2026groupmembench,owolabi2026socialmembench,hu2026evermembench}. The problem is therefore not merely finding relevant text, but preserving and recovering relations and historical structure in multi-party dialogue.

We organize this problem around two coupled challenges: \textbf{message attribution}, which distinguishes who said what, whom the content concerns, and whether information is personal or shared; and \textbf{state reconstruction}, which recovers current or historical states from clues distributed across members, groups, and time~\citep{ghosal2019dialoguegcn,li2020molweni,owolabi2026socialmembench,hu2026evermembench}. Global relevance retrieval does not guarantee that a low-frequency member or the correct branch enters a limited candidate set, while summaries, fact aggregation, and generic memory graphs may lose source/owner, PERSON/GROUP scope, or state versions~\citep{robertson2009bm25,karpukhin2020dpr,lewis2020rag,zhong2023memorybank,chhikara2025mem0,yue2026hypermem,xu2026structmem,gutierrez2024hipporag,tang2026mnemis}. Multi-party memory therefore needs verifiable evidence with participant relations and query-conditioned local state reconstruction, rather than simply more summaries or graph edges.

\begin{figure}[!t]
  \centering
  \includegraphics[width=\textwidth]{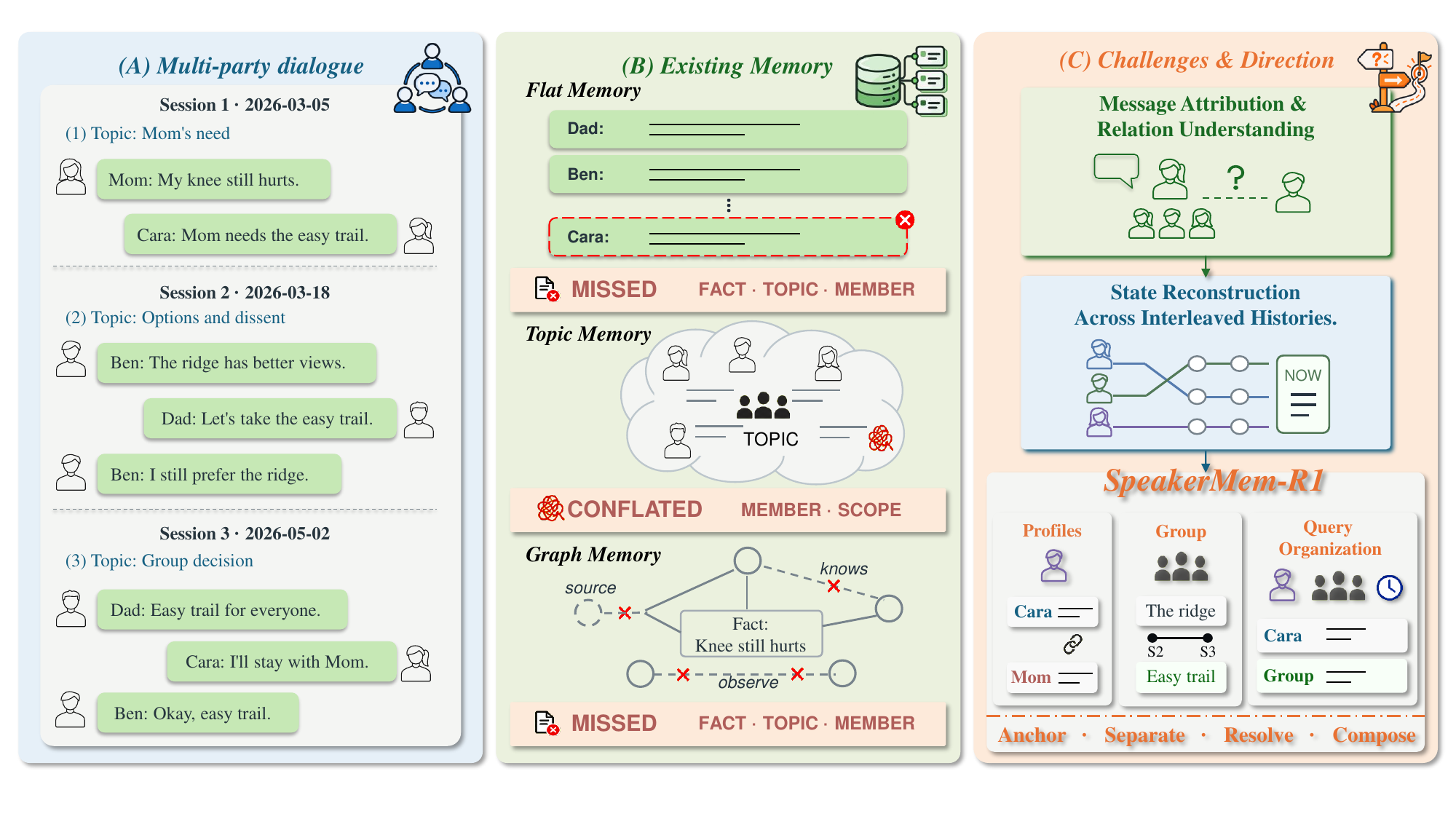}
  \caption{A multi-party group chat is not a flat message stream: participants discuss different topic branches, express different positions, refer to one another, and revise earlier states. Flat retrieval, aggregate summaries, and generic graph structures can lose these relations. \method{} preserves attribution and reconstructs state with dual-track memory and the Anchor--Separate--Resolve--Compose query procedure.}
  \label{fig:problem}
\end{figure}

We propose \method{}, a dual-track system that preserves speaker-labeled messages alongside provenance-linked derived states organized into person-level and group-level views. At query time, Anchor--Separate--Resolve--Compose retrieves and organizes evidence by participant, event, and time. We train a locally deployable Writer with SpeakerLevenshtein and speaker-conditioned GRPO to reduce attribution and update errors while freezing query and answer modules.

On GroupMemBench, SocialMemBench, and EverMemBench, \method{} achieves binary accuracies of 47.9\%, 69.2\%, and 61.9\%, respectively. Compared with the best results of mainstream frameworks evaluated on each benchmark, the scores are higher by 3.3, 12.4, and 9.4 percentage points, respectively. On the publicly reported EverMemBench leaderboard from EverMind-AI, \method{} achieves 62.33\%, the best reported result among the latest state-of-the-art frameworks. In a 305-question controlled evaluation under the main protocol, the Qwen2.5-3B Writer-R1 reaches 68.20\%, 10.82 points above SFT and within 3.28 points of the $71.48\%$ LLM writer reference; ablations confirm the complementary roles of both tracks and structured views.

Our contributions are as follows:
\begin{itemize}[leftmargin=*,nosep]
  \item We design \method{}, a dual-track memory system that combines traceable verbatim messages with person-level and group-level structured views for attribution, scope control, and state reconstruction in multi-party dialogue;
  \item We provide an analysis framework derived from question requirements and recurring error patterns in three multi-party benchmarks, covering member coverage, information attribution, personal/group scope, term and event disambiguation, and temporal updates with multi-hop reasoning;
  \item We train a locally deployable Qwen2.5-3B Writer with SpeakerLevenshtein and speaker-conditioned GRPO; on 305 controlled questions, RL writer reaches 95.4\% of the LLM writer reference accuracy, while the dual-track design and writing objective are evaluated on three multi-party benchmarks, LoCoMo, and targeted ablations.
\end{itemize}

\section{Related Work}
\label{sec:related}

\noindent\textbf{Long-term conversational memory and multi-party memory benchmarks.}
LoCoMo and LongMemEval evaluate factual, temporal, multi-hop, knowledge-update, and abstention abilities in long-term conversations~\citep{maharana2024locomo,wu2025longmemeval}; MemBench, MemoryAgentBench, StoryBench, and REALTALK extend this scope to reflective memory, test-time learning, dynamic branches, and real-world interaction~\citep{tan2025membench,hu2025memoryagentbench,wan2025storybench,lee2025realtalk}. In multi-party settings, DialogueGCN and Molweni establish the importance of speaker relations and discourse structure~\citep{ghosal2019dialoguegcn,li2020molweni}, while GroupMemBench, SocialMemBench, and EverMemBench evaluate group dynamics, social relations and norms, and cross-group collaboration with evolving states~\citep{yang2026groupmembench,owolabi2026socialmembench,hu2026evermembench}. These benchmarks show that group-chat evidence is distributed and updated across members, groups, and time rather than reducible to a flat message sequence.

\noindent\textbf{Retrieval augmentation, structured memory, and memory graphs.}
BM25, dense retrieval, and RAG/REALM/RETRO provide lexical or semantic matching but generally do not constrain member coverage, information attribution, or event consistency~\citep{robertson2009bm25,karpukhin2020dpr,lewis2020rag,guu2020realm,borgeaud2022retro}. MemoryBank, Mem0, A-MEM, MemGPT, generative agents, MemoBase, MemOS, and Zep study persistent facts, updates, connections, profiles, and temporal knowledge organization~\citep{zhong2023memorybank,chhikara2025mem0,xu2025amem,packer2023memgpt,park2023generative,memobase2024,li2025memos,rasmussen2025zep}, whereas MemoryLLM and M+ write memory into latent spaces~\citep{wang2024memoryllm,wang2025mplus}. EverMemOS/EverOS, RippleMem, MIRIX, HippoRAG, RAPTOR, GraphRAG, LightMem, StructMem, Mnemis, HyperMem, and LightRAG further use consolidation, event graphs, multi-agent collaboration, recursive summaries, and graph or hypergraph retrieval~\citep{hu2026evermemos,evermind2026multiround,ji2026ripplemem,wang2025mirix,gutierrez2024hipporag,sarthi2024raptor,edge2024graphrag,fang2025lightmem,xu2026structmem,tang2026mnemis,yue2026hypermem,guo2024lightrag}. These methods improve compression and association, but do not necessarily preserve key information in multi-party group chats, such as message attribution, cross-person cognition, and group consensus, among other aspects.

\noindent\textbf{Learnable memory management.}
Memory-R1, Mem-$\alpha$, Agentic Memory, DeltaMem, and Memory-R2 learn memory actions, hierarchical construction, multi-step management, state-difference rewards, or long-horizon credit assignment~\citep{yan2025memoryr1,wang2025memalpha,yu2026agentic,zhang2026deltamem,yan2026memoryr2}. CoMAM jointly optimizes memory agents~\citep{mao2026comam}, and DeferMem trains query-time evidence distillation~\citep{yin2026defermem}. In contrast, R$^2$-Mem uses RL-free reflective search~\citep{wang2026r2mem}; G-Memory organizes multi-agent experience in a hierarchy~\citep{zhang2025gmemory}; Reflexion uses verbal feedback without updating model weights~\citep{shinn2023reflexion}. Our focus is owner-level writing supervision in group chats, training only the Writer while retaining fixed retrieval and answering modules.

\section{Method}
\label{sec:method}

\begin{figure}[!t]
  \centering
  \includegraphics[width=\textwidth]{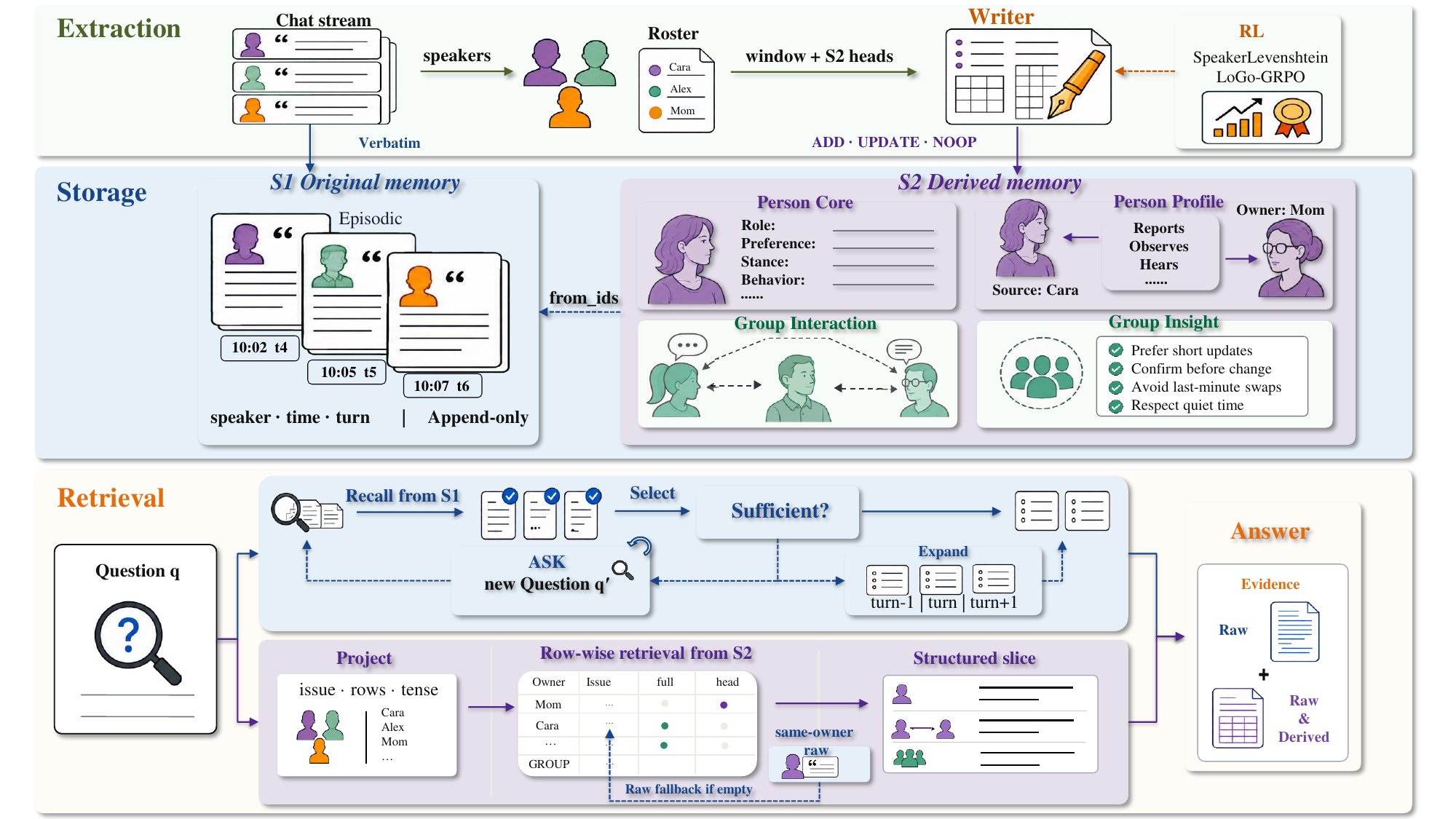}
  \caption{Overall architecture of \method{}. The system writes traceable System~1 verbatim memory and System~2 structures with participant, event, and state information, selects evidence from the two tracks at query time, and organizes it into the evidence set required for answering. RL rewards update only the Writer, while the answerer remains frozen.}
  \label{fig:architecture}
\end{figure}

The overall architecture is shown in Figure~\ref{fig:architecture}. Given a multi-party message stream with speaker, time, and channel information, \method{} first writes the messages into a traceable dual-track memory, then reconstructs query-specific evidence from the two tracks, and finally passes the evidence to a frozen answerer. The Writer is the component responsible for converting local messages into structured memory actions; the retrieval and answering stages operate on the resulting memory.

\subsection{Problem Definition: From Relevance Retrieval to Attributed Evidence}

Given a dialogue stream with $K\ge2$ participants,
\begin{equation}
D=\{u_t\}_{t=1}^{T},\qquad u_t=(x_t,s_t,\tau_t,c_t),
\end{equation}
where $x_t,s_t,\tau_t,c_t$ denote the text, speaker, time, and channel, respectively. The system first writes memory $M=W_\theta(D)$, then retrieves evidence and generates an answer with a frozen answerer:
\begin{equation}
E_q=R(q;M),\qquad \hat y=A(q,E_q).
\end{equation}
Ordinary retrieval optimizes only the relevance between evidence and $q$. Multi-party QA additionally requires evidence to be mutually compatible in participants, attribution, scope, event, and time. Because future questions are unknown at write time, a single summary or fixed event structure may fail to recover the evidence scope required by a query.

We represent a derived record as
\begin{equation}
r=(x,\mathrm{src},\mathrm{own},\mathrm{scope},\mathrm{event},\mathrm{time},\mathrm{state},\mathrm{ref}),
\label{eq:schema}
\end{equation}
where $x$ is the content, $\mathrm{src}$ and $\mathrm{own}$ are the source and owner, $\mathrm{scope}\in\{\textsc{Person},\textsc{Group}\}$ is the scope, and the final four fields denote event, time, state, and a source reference. A self-report typically satisfies $\mathrm{src}=\mathrm{own}$, whereas ``Alice believes that Bob has agreed'' satisfies $\mathrm{src}=\text{Alice},\mathrm{own}=\text{Bob}$. Thus, source--owner distinguishes who provides the information from whom the content concerns.

\paragraph{End-to-end execution.}

The system first writes memory and then answers queries. Messages enter System~1 verbatim, the Writer reads each local segment together with the roster and current System~2 state, and deterministic code validates the actions and adds provenance; at query time, Project produces query constraints, the two tracks retrieve and Compose combines evidence, and the frozen answerer produces the answer.

\paragraph{Online writing and non-destructive updates.}

System~1 appends messages without a language model and retains their source coordinates. The Writer reads each local segment and the current heads of the derived layers, returns \textsc{Add}, \textsc{Update}, or \textsc{Noop}, and deterministic code validates the action, adds provenance, and writes System~2. \textsc{Update} cites an existing \texttt{entry\_id}, appends a new node while reusing the owner, source, and layer coordinates of the record identified by that \texttt{entry\_id}, and connects the states through \texttt{links}/\texttt{superseded\_by} without overwriting history; the resulting chain supports head and full queries. Detailed constraints are given in Appendix~\ref{app:algorithms}.

\subsection{Five-Layer Traceable Dual-Track Memory}
\label{sec:five-layer-memory}

\method{} retains two complementary tracks. System~1 is the only verbatim layer and stores each message with its text, speaker, time, and channel. System~2 is a four-layer derived structure: PERSON-scoped Core (stable identity, facts, stances, and recurring behavior) and Profile (observations about a person or cross-person cognition), plus GROUP-scoped Interaction (cross-speaker events, relations, and decisions) and Insight (group norms, consensus, and exceptions). PERSON/GROUP fixes the record scope, source/owner separates who provides information from whom it concerns, and \texttt{from\_ids} links every derived record to supporting messages. The verbatim track therefore supplies exact wording and local context, while the derived track supplies person-level and group-level state views; the full fields, code-level layer names, and query roles are given in Appendix Table~\ref{tab:five-layers}.

\subsection{Query-Conditioned Evidence}
\label{sec:s1-s2-retrieval}

The five-layer store retains only information that can be recomposed; the final evidence set is query-dependent. Project compiles the query and deterministic roster into common query constraints:
\begin{equation}
Q_q=(\mathrm{rows}(q),\mathrm{issue}(q),\mathrm{mode}(q),\mathrm{scope}(q)).
\end{equation}
Here $\mathrm{rows}(q)$ contains the PERSON/GROUP rows to address, $\mathrm{issue}(q)$ is the issue or event constraint, $\mathrm{mode}(q)\in\{\mathrm{head},\mathrm{full}\}$ is the temporal mode, and $\mathrm{scope}(q)$ is the source--owner constraint. For example, ``every person'' expands all roster rows, ``the final decision'' selects GROUP rows, and ``Alice's view of Bob'' fixes source=Alice and owner=Bob.

\paragraph{Two retrieval paths and four-step organization.}
System~1 retrieves exact wording and local context from the verbatim track, optionally expands neighboring messages with Expand, and performs one supplementary search through Sufficiency/ASK when needed. System~2 first expands PERSON/GROUP rows according to $\mathrm{rows}(q)$, then selects records within each row by issue, relation, event, and time:
\begin{align}
C_p(q)&=\{r\in M^{\mathrm{der}}:\ \operatorname{own}(r)=p,\ \operatorname{match}(r,q)=1\},\\
V_q[p]
&=\operatorname{Temporal}_{\operatorname{mode}(q)}
  \big(\operatorname{Select}(C_p(q))\big),\qquad p\in\operatorname{rows}(q).
\label{eq:query-view}
\end{align}
The two retrieval paths use independent budgets. When a derived row is empty, the system falls back to the corresponding person's verbatim messages; if evidence is still unavailable, it preserves an explicit empty row. Candidate budgets and supplementary retrieval details are given in Appendix~\ref{app:implementation}. As an abstract summary of the system's behavior, \textbf{Anchor} retains source, owner, event, time, and source provenance; \textbf{Separate} expands PERSON/GROUP rows from the roster; \textbf{Resolve} distinguishes issues, parallel events, and current versus historical versions within each row; and \textbf{Compose} organizes the two evidence paths along persons, relations, and update chains before passing them to the frozen answerer. These four operations summarize the preceding write and query behavior rather than introducing an additional execution stage.

Appendix Table~\ref{tab:problem-mapping} maps five analysis dimensions to these mechanisms. The dimensions summarize benchmark question requirements and recurring error patterns, rather than independently annotated diagnostic labels.
\section{Writer Training Method}
\label{sec:r1-training}

At the system level, the Writer is model-agnostic; this section separately studies RL as a way to replace an expensive prompt-based Writer with a locally deployable small model. RL trains the \textsc{Add}/\textsc{Update}/\textsc{Noop} decisions of Qwen2.5-3B, while System~1 writing, query organization, and answering remain frozen. Local structural signals identify owner-level writing errors, while terminal QA gain measures their downstream effect after the conversation has been written.

\begin{figure}[!t]
  \centering
  \includegraphics[width=\textwidth]{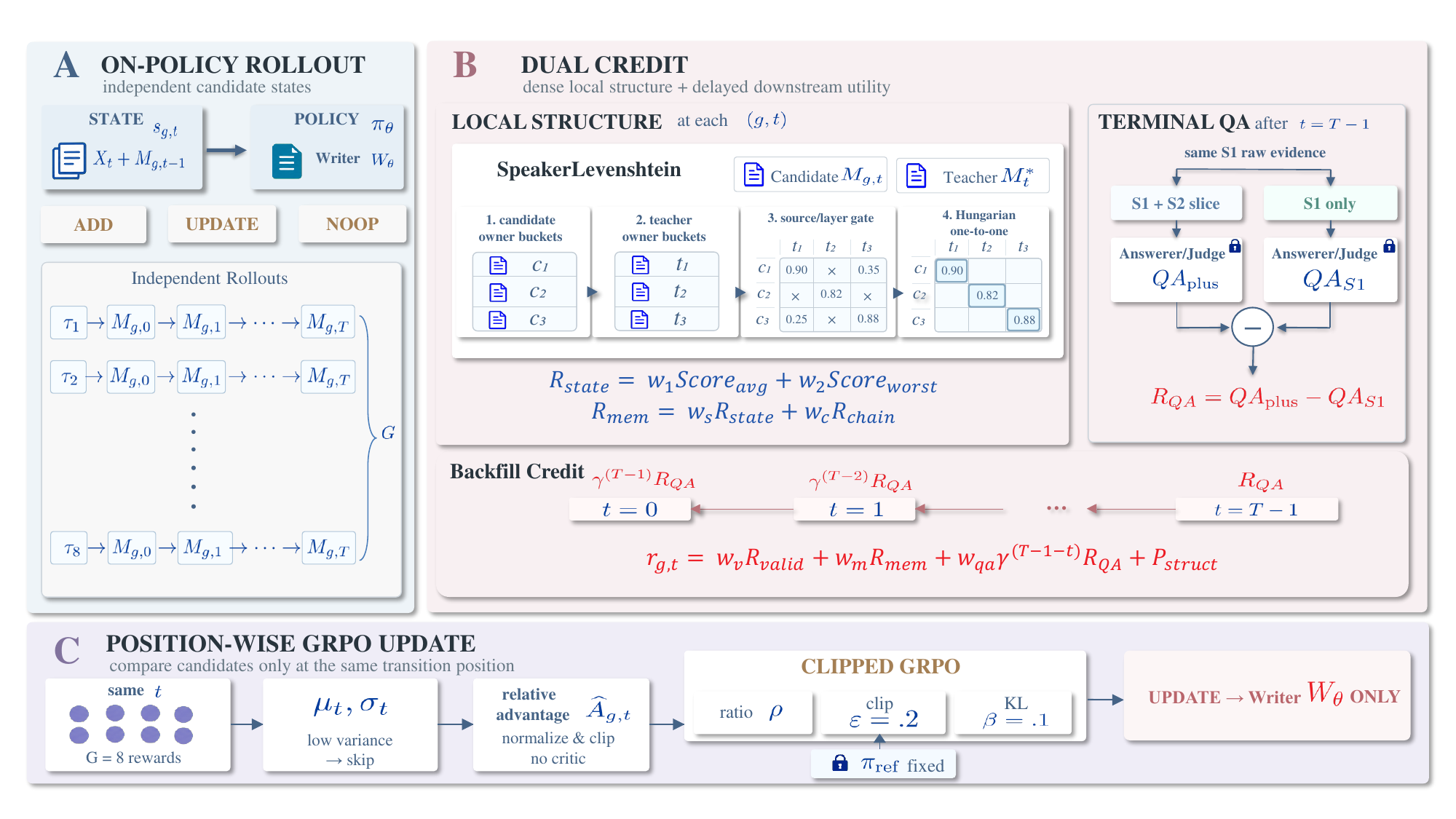}
  \caption{Speaker-Conditioned LoGo-GRPO training loop for Writer-R1. We sample $G=8$ writing trajectories for the same conversation. At each position, owner-decomposed SpeakerLevenshtein signals and terminal QA gains form group-relative advantages; clipped GRPO updates are applied only to the Writer, while retrieval and the answerer remain frozen.}
  \label{fig:rl-training}
\end{figure}

\paragraph{Evaluating structured states by owner.}
Inspired by DeltaMem's state-level memory matching~\citep{zhang2026deltamem}, \emph{SpeakerLevenshtein} combines token-level F1 with a normalized sequence-matching rate, rather than standard edit distance. It performs coordinate-consistent one-to-one matching within owner buckets: personal records, GROUP records, and cross-person observations cannot cancel one another, while omissions and over-writing are penalized. Let $P$ be the owner set and $F_p$ the matching result for owner $p$; the local structural potential is
\begin{equation}
\Phi_{\mathrm{SL}}(M,M^\star)
=w_1\frac{1}{|P|}\sum_{p\in P}F_p
+w_2\min_{p\in P}F_p.
\label{eq:speaker-levenshtein}
\end{equation}
The macro-average term measures the overall state, while the worst-owner term prevents frequent people from masking infrequent people or GROUP. We use $w_1=0.80$ and $w_2=0.20$. Detailed source/owner/layer gating, soft matching, and Hungarian alignment are given in Appendix~\ref{app:additional}.

\paragraph{Local-to-global returns.}
Memory-R2's LoGo-GRPO combines global optimization with local rerollouts from shared memory states~\citep{yan2026memoryr2}. Our speaker-conditioned variant instead compares aligned writing positions, using owner-level state scores and terminal QA gain. The local signal evaluates UPDATE transitions and structural validity. The global signal measures the gain of System~1+System~2 over System~1-only at $M_{g,T}$, under the same System~1 evidence:
\begin{equation}
R_g^{\mathrm{QA}}=\operatorname{QA}(\text{System~1}+\text{System~2};M_{g,T})
-\operatorname{QA}(\text{System~1};M_{g,T}).
\label{eq:terminal-qa-delta}
\end{equation}
This difference does not attribute questions already answerable from verbatim memory to the Writer. The return at writing position $t$ on trajectory $g$ is
\begin{equation}
r_{g,t}=w_{\mathrm{valid}}R^{\mathrm{valid}}_{g,t}+w_{\mathrm{mem}}R^{\mathrm{mem}}_{g,t}+P_{g,t}
+w_{\mathrm{QA}}\gamma^{T-1-t}R_g^{\mathrm{QA}},\qquad \gamma=0.95.
\label{eq:sc-grpo-return}
\end{equation}
We use $w_{\mathrm{valid}}=0.20$, $w_{\mathrm{mem}}=0.45$, and $w_{\mathrm{QA}}=0.35$. We sample multiple trajectories for the same network, compute group-relative advantages only at the same writing positions, and update the Writer with clipped GRPO~\citep{shao2024deepseekmath}. Positions with no effective within-group variation are omitted from the update. Reward decomposition, advantage computation, KL constraints, training data, and hyperparameters are given in Appendix~\ref{app:additional}.

\section{Experiments}
\label{sec:experiments}

\subsection{Research Questions and Evaluation Protocol}

We investigate whether dual-track memory improves multi-party QA, how individual components contribute to performance, and which types of errors remain.

\begin{table}[!t]
\caption{Main results (\%). GM/SM denote GroupMemBench/SocialMemBench; EM-MC/OE are EverMemBench subsets, while EM-All is the complete question set. Paired cells report Acc./token-F1; MeanQ weights questions and MeanN weights networks. The two model blocks test cross-model robustness across memory construction, retrieval, and answering under the evaluation protocol in Section~\ref{sec:experiments}. $^\dagger$: ASK disabled; Full context is feasible only on SM. Bold/underline mark the best/second-best non-Full-context result per block.}
\label{tab:main-results}
\centering\scriptsize
\renewcommand{\arraystretch}{1.04}
\setlength{\tabcolsep}{1.2pt}
\begin{tabularx}{\textwidth}{@{}l l *{7}{>{\centering\arraybackslash}X}@{}}
\toprule
\textbf{Model} & \textbf{Method} & \textbf{GM} & \textbf{SM} & \textbf{MeanQ} & \textbf{MeanN} & \textbf{EM-MC} & \textbf{EM-OE} & \textbf{EM-All} \\
\midrule
\multirow{8}{*}{\makecell{DeepSeek-V4-\\Flash}}
& BM25 & 44.6/25.0 & 28.6/15.7 & .234 & .250 & 64.4/3.6 & 27.0/\best{23.3} & 52.5/9.9 \\
& Embed & 34.5/17.4 & 38.1/18.1 & .300 & .307 & 56.1/4.0 & 23.9/\second{22.7} & 45.9/10.0 \\
& Mem0 & 21.6/9.0 & 13.7/11.0 & .132 & .130 & 28.4/5.0 & 1.7/7.0 & 19.9/6.0 \\
& A-MEM & 27.1/10.6 & 56.8/\second{31.9} & .608 & .610 & 37.7/14.2 & 8.8/12.9 & 28.5/13.8 \\
& HippoRAG & 27.0/11.1 & 55.9/30.6 & .584 & .574 & 62.9/18.5 & 15.9/18.4 & 48.0/18.5 \\
& Full context & --- & 69.4/26.1 & .592 & .573 & --- & --- & --- \\
\cmidrule(lr){2-9}
\rowcolor{gray!10}
\cellcolor{white} & \method{}$^\dagger$ & \second{47.0}/\second{25.2} & \best{69.2}/27.4 & \best{.713} & \second{.691} & \second{71.2}/\second{33.7} & 37.4/16.7 & \second{60.5}/\best{28.3} \\
\rowcolor{gray!10}
\cellcolor{white} & \method{} & \best{47.9}/\best{26.5} & \second{64.9}/\best{32.7} & \second{.710} & \best{.693} & \best{72.0}/\best{33.8} & \best{40.0}/15.7 & \best{61.9}/\second{28.1} \\
\midrule
\multirow{7}{*}{\makecell{GPT-5.6-\\luna}}
& BM25 & \best{46.2}/\second{26.6} & 30.0/21.1 & .412 & .452 & 69.6/33.0 & \second{27.8}/\best{24.4} & 56.3/\second{30.3} \\
& Embed & 35.3/18.4 & 36.4/22.7 & .519 & .538 & 58.5/28.8 & 24.5/\second{23.3} & 47.8/27.1 \\
& Mem0 & 20.27/8.1 & 13.97/14.9 & .252 & .262 & 35.0/18.7 & 1.4/9.6 & 24.38/15.8 \\
& A-MEM & 26.58/9.8 & 46.56/\second{24.9} & .607 & .601 & 53.4/27.4 & 11.8/16.9 & 40.17/24.1 \\
& HippoRAG & 27.11/11.9 & \second{61.30}/\best{28.0} & \second{.726} & \best{.723} & \best{75.1}/\second{35.2} & 20.1/21.9 & \second{57.62}/\best{30.9} \\
& Full context & --- & 71.3/29.6 & .769 & .750 & --- & --- & --- \\
\cmidrule(lr){2-9}
\rowcolor{gray!10}
\cellcolor{white} & \method{} & \second{42.7}/\best{26.8} & \best{64.4}/24.1 & \best{.731} & \second{.715} & \second{71.3}/\best{35.6} & \best{35.8}/16.4 & \best{60.0}/29.5 \\
\bottomrule
\end{tabularx}
\end{table}

We evaluate GroupMemBench (745 questions), SocialMemBench (1,031), and EverMemBench (2,400)~\citep{yang2026groupmembench,owolabi2026socialmembench,hu2026evermembench} against BM25, dense retrieval, Mem0, A-MEM, HippoRAG, and Full context when feasible~\citep{chhikara2025mem0,xu2025amem,gutierrez2024hipporag}; LoCoMo (1,986 questions) serves as a two-person long-term conversation boundary test. Evaluation proceeds through memory construction, question-conditioned retrieval, and answer generation. The DeepSeek-V4-Flash and GPT-5.6-luna configurations switch the language model across these stages to assess cross-model robustness. For the primary three-benchmark comparison, each system follows its official code, recommended configuration, and official prompt; we standardize only the data, metric definitions, judge model, and evaluation interface, while retaining speaker/source metadata where supported. The public EverMemBench comparison uses a separate configuration specified in Table~\ref{tab:evermem-cross-config}. Our primary metric is question-level binary accuracy (Acc.; Appendix Eq.~\eqref{eq:binary-accuracy}), which measures task-level correctness; supplementary token-F1 (Appendix Eq.~\eqref{eq:token-f1}) measures judge-independent token overlap. SocialMem additionally reports MeanQ/MeanN (Appendix Eq.~\eqref{eq:socialmem-means}), which aggregate official rubric scores with equal weights for questions and networks, respectively. Acc. and token-F1 are reported as percentages, whereas MeanQ/MeanN lie in $[0,1]$; execution settings and category results appear in Appendices~\ref{app:benchmarks}--\ref{app:reproducibility} and~\ref{app:multibench-detailed}.

\subsection{Main Multi-Party Results}

Table~\ref{tab:main-results} shows highest \method{} accuracies of 47.9\%, 69.2\%, and 61.9\% on GroupMem, SocialMem, and EverMem, respectively, when selecting the highest-accuracy \method{} configuration for each benchmark. Among the non-Full-context mainstream baselines on each benchmark, the best results are 44.6\% from BM25, 56.8\% from A-MEM, and 52.5\% from BM25, respectively. Compared with the best results of mainstream frameworks evaluated on each benchmark, the scores are higher by 3.3, 12.4, and 9.4 percentage points, respectively. Full context is feasible only on SocialMemBench; the other histories cannot reliably fit within the configured context window. On SocialMem, the 69.2\% result is near Full context at 69.4\%; the full system obtains MeanQ/MeanN 0.710/0.693 with a network-level 95\% CI of $[0.659,0.726]$. The small MeanQ/MeanN reversal between the two variants reflects their different weighting of questions and networks.

\begin{figure}[!t]
  \centering
  \includegraphics[width=\linewidth]{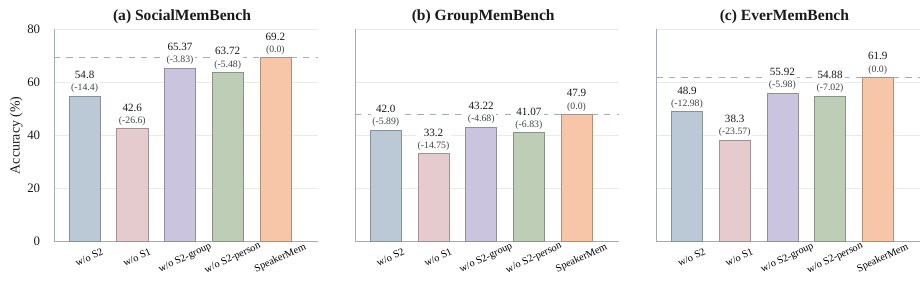}
  \caption{Memory-path and hierarchy ablations. Labels show accuracy and change from the best full dual-track result. Here, w/o S2 removes all four System~2 derived layers, w/o S1 removes the verbatim track, w/o S2-group removes the two GROUP layers (Interaction and Insight), and w/o S2-person removes the two PERSON layers (Core and Profile). SpeakerMem denotes the full dual-track system.}
  \label{fig:ablations}
\end{figure}

\begin{table}[!t]
\caption{EverMemBench accuracy (\%) across nine public behavior labels and the question-weighted total over 2,400 questions. All methods use GPT-4.1-mini for answering and Gemini-3-Flash for judging; \method{} uses System~1 top-10 and System~2 owner-2 + source-1. Bold/underline indicate best/second-best.}
\label{tab:evermem-cross-config}
\centering
\fontsize{7.4}{8.6}\selectfont
\setlength{\tabcolsep}{1.0pt}
\begin{tabularx}{\textwidth}{@{}l*{10}{>{\centering\arraybackslash}X}@{}}
\toprule
\textbf{Method} & \textbf{Single} & \textbf{Multi} & \textbf{Temp} & \textbf{Const} & \textbf{Proact} & \textbf{Update} & \textbf{Style} & \textbf{Skill} & \textbf{Role} & \makecell{\textbf{Weighted}\\\textbf{total}} \\
\midrule
MemoBase & 60.09 & 12.85 & 18.00 & 64.68 & 36.77 & 30.60 & 17.05 & 29.59 & 38.78 & 36.21 \\
Mem0 & 55.40 & 11.24 & 6.33 & 66.17 & 52.46 & 51.87 & 22.73 & 31.36 & 36.22 & 39.92 \\
Zep & 73.71 & 8.03 & 13.00 & 67.16 & 47.54 & 43.66 & 26.70 & 35.50 & 44.39 & 41.67 \\
MemOS & 71.36 & 18.88 & 15.67 & 69.90 & 51.99 & 45.15 & 28.98 & 32.54 & 48.47 & 44.63 \\
RippleMem & 92.02 & 22.09 & \second{21.33} & 78.11 & \second{71.66} & 58.96 & 31.25 & 36.69 & \best{53.06} & 54.75 \\
EverOS & \best{94.37} & \best{28.11} & 20.33 & \second{86.07} & 68.62 & \best{84.70} & \second{39.77} & \best{42.60} & \second{52.04} & \second{60.08} \\
\midrule
\rowcolor{gray!10}
\method{} & \second{93.43} & \second{24.10} & \best{35.00} & \best{87.06} & \best{75.64} & \second{80.22} & \best{50.57} & \second{40.83} & 43.88 & \best{62.33} \\
\bottomrule
\end{tabularx}
\end{table}

\paragraph{Publicly reported EverMemBench leaderboard from EverMind-AI.}
On the publicly reported EverMemBench leaderboard from EverMind-AI, Table~\ref{tab:evermem-cross-config} compares all 2,400 EverMemBench questions using the GPT-4.1-mini/Gemini-3-Flash configuration~\citep{evermind2026multiround}. \method{} answers 1,496/2,400 questions correctly (62.33\% accuracy), versus approximately 60.08\% for EverOS and 54.75\% for RippleMem; public totals are reconstructed from rounded category results, whereas ours uses question-level records. Strong Single, Const, Proact, and Update results contrast with weaker Multi, Skill, and Role scores, exposing cross-evidence, preference, and role-attribution bottlenecks.

The accuracy gains depend on the task. On GroupMem, BM25, the global lexical-retrieval baseline, already reaches 44.6\%, while \method{} obtains 47.9\%. On SocialMem, A-MEM is the strongest mainstream baseline at 56.8\%, followed by HippoRAG at 55.9\%; the highest \method{} accuracy shown reaches 69.2\%. On EverMem open-ended questions, BM25 is the strongest comparison at 27.0\%, whereas \method{} achieves 40.0\%, a 13.0-point gain. Binary accuracy and token-F1 are not monotonic: the full system has lower SocialMem accuracy than $^\dagger$ but higher token-F1 (32.7 versus 27.4), because extra people or incorrect scope can invalidate an otherwise overlapping answer. We therefore report both.

\vspace{-1.2\baselineskip}
\subsection{Dual-Track and Hierarchical Ablations}
\vspace{-0.7\baselineskip}

Figure~\ref{fig:ablations} compares path and hierarchy ablations against the better full-system result with or without ASK. Removing either track reduces accuracy on all three benchmarks; retaining only per-speaker or group records also degrades performance, showing complementary evidence at both levels. Exact results appear in Appendix Table~\ref{tab:ablations}.

With ASK disabled, \method{}$^*$ obtains accuracies of 47.0/69.2/60.5\% on GroupMem/SocialMem/EverMem, versus 47.9/64.9/61.9\% with ASK. Supplementary retrieval recovers dispersed clues on GroupMem and EverMem but can add redundant evidence on SocialMem.

We additionally analyze joint S1/S2 retrieval top-$k$ sensitivity, with response curves, a dual-metric grid, and detailed discussion in Appendix~\ref{app:topk-sensitivity} (Figures~\ref{fig:topk-sensitivity-curves} and~\ref{fig:topk-sensitivity-grid}).

\begin{table}[!t]
\caption{Controlled Writer results on 10 held-out complete SocialMem networks (305 questions). All rows use the same no-ASK query protocol: System~1 recalls $n=40$ candidates and retains final top-$k=10$; System~2 uses $k=2$/source-$k=1$. Query and answer modules are frozen; Acc. reports mean$\pm$sample standard deviation over three runs.}
\label{tab:rl-results}
\centering\small
\renewcommand{\arraystretch}{1.05}
\setlength{\tabcolsep}{0.8pt}
\begin{tabularx}{\textwidth}{@{}>{\raggedright\arraybackslash}p{1.13in}>{\centering\arraybackslash}p{1.10in}>{\centering\arraybackslash}p{0.78in}>{\centering\arraybackslash}p{1.18in}>{\centering\arraybackslash}X@{}}
\toprule
\textbf{Writing strategy} & \textbf{Writer} & \textbf{Correct/total} & \textbf{Acc. (\%, mean$\pm$std.)} & \textbf{Gap to LLM writer} \\
\midrule
SFT (epoch 10) & Qwen2.5-3B & 175/305 & $57.38\pm0.33$ & $-14.10$ pp \\
\rowcolor{gray!10}
\textbf{Writer-R1 (30 steps)} & Qwen2.5-3B & 208/305 & $\mathbf{68.20\pm0.66}$ & $-3.28$ pp \\
LLM writer & DeepSeek-V4-Flash & 218/305 & $71.48\pm0.66$ & -- \\
\bottomrule
\end{tabularx}
\end{table}

\begin{table}[!t]
\caption{LoCoMo category accuracy (\%). The metric is LLM-as-a-judge accuracy (\%) scored by GPT-4o-mini. LightRAG uses its official implementation~\citep{guo2024lightrag}. Mem0, A-MEM, and Zep scores come from Tables 1--2 of \citet{chhikara2025mem0}; MemOS scores come from Table 3 of \citet{li2025memos_system}. ALL(Non-AD) aggregates the four non-AD categories. Bold/underline mark best/second-best.}
\label{tab:locomo-compatibility}
\centering\small
\renewcommand{\arraystretch}{1.05}
\setlength{\tabcolsep}{3pt}
\begin{tabularx}{\textwidth}{@{}l>{\centering\arraybackslash}X>{\centering\arraybackslash}X>{\centering\arraybackslash}X>{\centering\arraybackslash}X>{\centering\arraybackslash}X@{}}
\toprule
\textbf{Method} & \textbf{Single-hop} & \textbf{Multi-hop} & \textbf{Temporal} & \textbf{Open-domain} & \textbf{ALL(Non-AD)} \\
\midrule
Mem0 & 67.13 & 51.15 & 55.51 & \second{72.93} & 66.88 \\
A-MEM & 39.79 & 18.85 & 49.91 & 54.05 & 48.38 \\
MemOS & \second{81.09} & \second{67.49} & \best{75.18} & 55.90 & \second{75.80} \\
Zep & 61.70 & 41.35 & 49.31 & \best{76.60} & 65.99 \\
LightRAG & \best{86.68} & \best{84.04} & 60.75 & 71.88 & \best{79.87} \\
\midrule
\rowcolor{gray!10}
\method{} & 77.88 & 41.13 & \second{70.72} & 40.62 & 67.34 \\
\bottomrule
\end{tabularx}
\end{table}

\vspace{-0.8\baselineskip}
\subsection{Does the Full R1 Objective Improve Writing?}

On 10 held-out SocialMem networks (305 questions), using the main no-ASK query configuration (System~1 recall-$n=40$ with final top-$k=10$, System~2 $k=2$/source-$k=1$), the Qwen2.5-3B Writer-R1 reaches $68.20\pm0.66\%$ with the query and answer modules frozen, up from $57.38\pm0.33\%$ for SFT: a gain of 10.82 percentage points, or 33 additional correct answers (Table~\ref{tab:rl-results}). Under the same protocol, the LLM writer reference reaches $71.48\pm0.66\%$ (218/305), leaving R1 3.28 points behind and at 95.4\% of the LLM writer accuracy. This result shows that RL brings a locally deployable small Writer close to the LLM writer reference, while it is not evidence of broad cross-domain RL generalization. Training-data composition, hyperparameters, and per-seed results are reported in Appendix~\ref{app:additional}.

\paragraph{LoCoMo boundary test.} \method{} achieves an accuracy of 70.85\% (1,407/1,986) on two-person LoCoMo (Table~\ref{tab:locomo-compatibility}); multi-hop and open-domain questions remain weak. Full counts appear in Appendix~\ref{app:locomo-detailed}, and this test does not replace multi-party validation.

\section{Conclusion}

\method{} uses a simple, efficient, and traceable dual-track memory to address the attribution and state-reconstruction failures that general-purpose memory systems exhibit in long-term multi-party conversations. It preserves source-linked verbatim evidence, organizes derived states into person-level and group-level views, and combines the two tracks through constrained query-time composition. This design directly targets multi-party memory failure modes without relying on a complex collection of additional mechanisms. Three multi-party benchmarks and LoCoMo demonstrate the design's effectiveness and limits; ablations establish complementary contributions from the two tracks and person-level and group-level views. In the 305-question controlled study under the main protocol, the RL Writer (30 steps) reaches 95.4\% of the $71.48\%$ LLM writer reference accuracy. This result shows that our RL method can make a locally deployable small Writer approach the performance of the large Writer while keeping the memory pipeline practical. Broader RL generalization, full-member coverage, cross-evidence reasoning, and knowledge outside memory remain open challenges.

\paragraph{Limitations and future work.} The system assumes reliable rosters, source/owner attribution, and temporal identification; aliases, membership changes, implicit audiences, and parallel events remain challenging. Future work should reduce construction and retrieval cost and improve cross-evidence reasoning, open-domain QA, and generalization across domains and languages.

\clearpage
\bibliography{references/references}

\begin{thebibliography}{50}
\providecommand{\natexlab}[1]{#1}
\providecommand{\url}[1]{\texttt{#1}}
\expandafter\ifx\csname urlstyle\endcsname\relax
  \providecommand{\doi}[1]{doi: #1}\else
  \providecommand{\doi}{doi: \begingroup \urlstyle{rm}\Url}\fi

\bibitem[Borgeaud et~al.(2022)Borgeaud, Mensch, Hoffmann, Cai, Rutherford,
  Millican, van~den Driessche, Lespiau, Damoc, Clark, de~Las~Casas, Guy,
  Menick, Ring, Hennigan, Huang, Maggiore, Jones, Cassirer, Brock, Paganini,
  Irving, Vinyals, Osindero, Simonyan, Rae, Elsen, and
  Sifre]{borgeaud2022retro}
Sebastian Borgeaud, Arthur Mensch, Jordan Hoffmann, Trevor Cai, Eliza
  Rutherford, Katie Millican, George van~den Driessche, Jean-Baptiste Lespiau,
  Bogdan Damoc, Aidan Clark, Diego de~Las~Casas, Aurelia Guy, Jacob Menick,
  Roman Ring, Tom Hennigan, Saffron Huang, Loren Maggiore, Chris Jones, Albin
  Cassirer, Andy Brock, Michela Paganini, Geoffrey Irving, Oriol Vinyals, Simon
  Osindero, Karen Simonyan, Jack~W. Rae, Erich Elsen, and Laurent Sifre.
\newblock Improving language models by retrieving from trillions of tokens.
\newblock In \emph{Proceedings of the 39th International Conference on Machine
  Learning}, volume 162, pp.\  2206--2240, Baltimore, Maryland, USA, 2022.
  PMLR.
\newblock URL \url{https://proceedings.mlr.press/v162/borgeaud22a.html}.

\bibitem[Chhikara et~al.(2025)Chhikara, Khant, Aryan, Singh, and
  Yadav]{chhikara2025mem0}
Prateek Chhikara, Dev Khant, Saket Aryan, Taranjeet Singh, and Deshraj Yadav.
\newblock {Mem0}: Building production-ready {AI} agents with scalable long-term
  memory.
\newblock In \emph{Proceedings of the 27th European Conference on Artificial
  Intelligence}, pp.\  2993--3000, Bologna, Italy, 2025. IOS Press.
\newblock \doi{10.3233/FAIA251160}.

\bibitem[Edge et~al.(2024)Edge, Trinh, Cheng, Bradley, Chao, Mody, Truitt,
  Metropolitansky, Ness, and Larson]{edge2024graphrag}
Darren Edge, Ha~Trinh, Newman Cheng, Joshua Bradley, Alex Chao, Apurva Mody,
  Steven Truitt, Dasha Metropolitansky, Robert~Osazuwa Ness, and Jonathan
  Larson.
\newblock From local to global: A graph {RAG} approach to query-focused
  summarization.
\newblock arXiv preprint arXiv:2404.16130, 2024.
\newblock URL \url{https://arxiv.org/abs/2404.16130}.

\bibitem[{EverMind researchers}(2026)]{evermind2026multiround}
{EverMind researchers}.
\newblock Multi-round retrieval: Letting the model decide when to stop
  searching.
\newblock EverMind Blog, September 4 2026.
\newblock URL
  \url{https://evermind.ai/blogs/multi-round-retrieval-letting-the-model-decide-when-to-stop-searching}.
\newblock Accessed September 21, 2026.

\bibitem[Fang et~al.(2026)Fang, Deng, Xu, Jiang, Tang, Xu, Deng, Yao, Wang,
  Qiao, Chen, and Zhang]{fang2025lightmem}
Jizhan Fang, Xinle Deng, Haoming Xu, Ziyan Jiang, Yuqi Tang, Ziwen Xu, Shumin
  Deng, Yunzhi Yao, Mengru Wang, Shuofei Qiao, Huajun Chen, and Ningyu Zhang.
\newblock {LightMem}: Lightweight and efficient memory-augmented generation.
\newblock In \emph{Proceedings of the International Conference on Learning
  Representations}, 2026.

\bibitem[Ghosal et~al.(2019)Ghosal, Majumder, Poria, Chhaya, and
  Gelbukh]{ghosal2019dialoguegcn}
Deepanway Ghosal, Navonil Majumder, Soujanya Poria, Niyati Chhaya, and
  Alexander Gelbukh.
\newblock {DialogueGCN}: A graph convolutional neural network for emotion
  recognition in conversation.
\newblock In \emph{Proceedings of the 2019 Conference on Empirical Methods in
  Natural Language Processing and the 9th International Joint Conference on
  Natural Language Processing}, pp.\  154--164, Hong Kong, China, 2019.
  Association for Computational Linguistics.
\newblock \doi{10.18653/v1/D19-1015}.

\bibitem[Guo et~al.(2024)Guo, Xia, Yu, Ao, and Huang]{guo2024lightrag}
Zirui Guo, Lianghao Xia, Yanhua Yu, Tu~Ao, and Chao Huang.
\newblock {LightRAG}: Simple and fast retrieval-augmented generation.
\newblock arXiv preprint arXiv:2410.05779, 2024.
\newblock URL \url{https://arxiv.org/abs/2410.05779}.

\bibitem[Guti{\'e}rrez et~al.(2024)Guti{\'e}rrez, Shu, Gu, Yasunaga, and
  Su]{gutierrez2024hipporag}
Bernal~Jim{\'e}nez Guti{\'e}rrez, Yiheng Shu, Yu~Gu, Michihiro Yasunaga, and
  Yu~Su.
\newblock {HippoRAG}: Neurobiologically inspired long-term memory for large
  language models.
\newblock In \emph{Advances in Neural Information Processing Systems 37}, pp.\
  59532--59569, Vancouver, Canada, 2024. Neural Information Processing Systems
  Foundation.
\newblock \doi{10.52202/079017-1902}.
\newblock URL
  \url{https://proceedings.nips.cc/paper_files/paper/2024/hash/6ddc001d07ca4f319af96a3024f6dbd1-Abstract-Conference.html}.

\bibitem[Guu et~al.(2020)Guu, Lee, Tung, Pasupat, and Chang]{guu2020realm}
Kelvin Guu, Kenton Lee, Zora Tung, Panupong Pasupat, and Ming-Wei Chang.
\newblock Retrieval augmented language model pre-training.
\newblock In \emph{Proceedings of the 37th International Conference on Machine
  Learning}, volume 119, pp.\  3929--3938, Virtual, 2020. PMLR.
\newblock URL \url{https://proceedings.mlr.press/v119/guu20a.html}.

\bibitem[Hu et~al.(2026{\natexlab{a}})Hu, Gao, Zhou, Xu, Bai, Li, Zhang, Li,
  Zhang, Bing, and Deng]{hu2026evermemos}
Chuanrui Hu, Xingze Gao, Zuyi Zhou, Dannong Xu, Yi~Bai, Xintong Li, Hui Zhang,
  Tong Li, Chong Zhang, Lidong Bing, and Yafeng Deng.
\newblock {EverMemOS}: A self-organizing memory operating system for structured
  long-horizon reasoning.
\newblock In \emph{Proceedings of the 64th Annual Meeting of the Association
  for Computational Linguistics (ACL 2026)}, pp.\  45836--45853, San Diego,
  California, USA, 2026{\natexlab{a}}. Association for Computational
  Linguistics.
\newblock \doi{10.18653/v1/2026.acl-long.2125}.
\newblock URL \url{https://aclanthology.org/2026.acl-long.2125/}.

\bibitem[Hu et~al.(2026{\natexlab{b}})Hu, Li, Gao, Chen, Bai, Xu, Lin, Li, Han,
  Pei, and Deng]{hu2026evermembench}
Chuanrui Hu, Tong Li, Xingze Gao, Hongda Chen, Yi~Bai, Dannong Xu, Tianwei Lin,
  Xiaohong Li, Yunyun Han, Jian Pei, and Yafeng Deng.
\newblock Evaluating long-horizon memory for multi-party collaborative
  dialogues.
\newblock In \emph{Proceedings of the 32nd ACM SIGKDD Conference on Knowledge
  Discovery and Data Mining V.2}, pp.\  9082--9090. Association for Computing
  Machinery, 2026{\natexlab{b}}.
\newblock \doi{10.1145/3770855.3817589}.

\bibitem[Hu et~al.(2025)Hu, Wang, and McAuley]{hu2025memoryagentbench}
Yuanzhe Hu, Yu~Wang, and Julian McAuley.
\newblock Evaluating memory in {LLM} agents via incremental multi-turn
  interactions.
\newblock arXiv preprint arXiv:2507.05257, 2025.
\newblock URL \url{https://arxiv.org/abs/2507.05257}.

\bibitem[Ji et~al.(2026)Ji, Li, Cheng, Zhou, Zhang, Tan, and
  Qin]{ji2026ripplemem}
Jingbo Ji, Lingyi Li, Xilong Cheng, Yuhao Zhou, Wenji Zhang, Yuting Tan, and
  Yunxiao Qin.
\newblock {RippleMem}: From isolated retrieval to associative recollection for
  long-term agent memory.
\newblock arXiv preprint arXiv:2608.13334, 2026.
\newblock URL \url{https://arxiv.org/abs/2608.13334}.

\bibitem[Karpukhin et~al.(2020)Karpukhin, Oguz, Min, Lewis, Wu, Edunov, Chen,
  and Yih]{karpukhin2020dpr}
Vladimir Karpukhin, Barlas Oguz, Sewon Min, Patrick Lewis, Ledell Wu, Sergey
  Edunov, Danqi Chen, and Wen-tau Yih.
\newblock Dense passage retrieval for open-domain question answering.
\newblock In \emph{Proceedings of the 2020 Conference on Empirical Methods in
  Natural Language Processing}, pp.\  6769--6781, Online, 2020. Association for
  Computational Linguistics.
\newblock \doi{10.18653/v1/2020.emnlp-main.550}.

\bibitem[Lee et~al.(2025)Lee, Maharana, Pujara, Ren, and
  Barbieri]{lee2025realtalk}
Dong-Ho Lee, Adyasha Maharana, Jay Pujara, Xiang Ren, and Francesco Barbieri.
\newblock {REALTALK}: A 21-day real-world dataset for long-term conversation.
\newblock arXiv preprint arXiv:2502.13270, 2025.
\newblock URL \url{https://arxiv.org/abs/2502.13270}.

\bibitem[Lewis et~al.(2020)Lewis, Perez, Piktus, Petroni, Karpukhin, Goyal,
  K{\"u}ttler, Lewis, Yih, Rockt{\"a}schel, Riedel, and Kiela]{lewis2020rag}
Patrick Lewis, Ethan Perez, Aleksandra Piktus, Fabio Petroni, Vladimir
  Karpukhin, Naman Goyal, Heinrich K{\"u}ttler, Mike Lewis, Wen-tau Yih, Tim
  Rockt{\"a}schel, Sebastian Riedel, and Douwe Kiela.
\newblock Retrieval-augmented generation for knowledge-intensive {NLP} tasks.
\newblock In \emph{Advances in Neural Information Processing Systems 33}, pp.\
  9459--9474, Vancouver, Canada, 2020. Neural Information Processing Systems
  Foundation.
\newblock URL
  \url{https://papers.nips.cc/paper/2020/hash/6b493230205f780e1bc26945df7481e5-Abstract.html}.

\bibitem[Li et~al.(2020)Li, Liu, Kan, Zheng, Wang, Lei, Liu, and
  Qin]{li2020molweni}
Jiaqi Li, Ming Liu, Min-Yen Kan, Zihao Zheng, Zekun Wang, Wenqiang Lei, Ting
  Liu, and Bing Qin.
\newblock {Molweni}: A challenge multiparty dialogue-based machine reading
  comprehension dataset with discourse structure.
\newblock In \emph{Proceedings of the 28th International Conference on
  Computational Linguistics}, pp.\  2642--2652, Barcelona, Spain, 2020.
  International Committee on Computational Linguistics.
\newblock \doi{10.18653/v1/2020.coling-main.238}.
\newblock URL \url{https://aclanthology.org/2020.coling-main.238/}.

\bibitem[Li et~al.(2025{\natexlab{a}})Li, Song, Wang, Niu, Chen, Yang, Xi, Lai,
  Zhao, Wang, Ren, Lin, Huo, Chen, Chen, Li, Yin, Yu, Tang, Yang, Xu, and
  Xiong]{li2025memos}
Zhiyu Li, Shichao Song, Hanyu Wang, Simin Niu, Ding Chen, Jiawei Yang, Chenyang
  Xi, Huayi Lai, Jihao Zhao, Yezhaohui Wang, Junpeng Ren, Zehao Lin, Jiahao
  Huo, Tianyi Chen, Kai Chen, Kehang Li, Zhiqiang Yin, Qingchen Yu, Bo~Tang,
  Hongkang Yang, Zhi-Qin~John Xu, and Feiyu Xiong.
\newblock {MemOS}: An operating system for memory-augmented generation ({MAG})
  in large language models.
\newblock arXiv preprint arXiv:2505.22101, 2025{\natexlab{a}}.
\newblock URL \url{https://arxiv.org/abs/2505.22101}.

\bibitem[Li et~al.(2025{\natexlab{b}})Li, Xi, Li, Chen, Chen, Song, Niu, Wang,
  Yang, Tang, Yu, Zhao, Wang, Liu, Lin, Wang, Huo, Chen, Chen, Li, Tao, Lai,
  Wu, Tang, Wang, Fan, Zhang, Zhang, Yan, Yang, Xu, Xu, Chen, Wang, Yang,
  Zhang, Xu, Chen, and Xiong]{li2025memos_system}
Zhiyu Li, Chenyang Xi, Chunyu Li, Ding Chen, Boyu Chen, Shichao Song, Simin
  Niu, Hanyu Wang, Jiawei Yang, Chen Tang, Qingchen Yu, Jihao Zhao, Yezhaohui
  Wang, Peng Liu, Zehao Lin, Pengyuan Wang, Jiahao Huo, Tianyi Chen, Kai Chen,
  Kehang Li, Zhen Tao, Huayi Lai, Hao Wu, Bo~Tang, Zhengren Wang, Zhaoxin Fan,
  Ningyu Zhang, Linfeng Zhang, Junchi Yan, Mingchuan Yang, Tong Xu, Wei Xu,
  Huajun Chen, Haofen Wang, Hongkang Yang, Wentao Zhang, Zhi-Qin~John Xu,
  Siheng Chen, and Feiyu Xiong.
\newblock {MemOS}: A memory {OS} for {AI} system.
\newblock arXiv preprint arXiv:2507.03724v4, 2025{\natexlab{b}}.
\newblock URL \url{https://arxiv.org/abs/2507.03724v4}.

\bibitem[Maharana et~al.(2024)Maharana, Lee, Tulyakov, Bansal, Barbieri, and
  Fang]{maharana2024locomo}
Adyasha Maharana, Dong-Ho Lee, Sergey Tulyakov, Mohit Bansal, Francesco
  Barbieri, and Yuwei Fang.
\newblock Evaluating very long-term conversational memory of {LLM} agents.
\newblock In \emph{Proceedings of the 62nd Annual Meeting of the Association
  for Computational Linguistics (Volume 1: Long Papers)}, pp.\  13851--13870.
  Association for Computational Linguistics, 2024.
\newblock \doi{10.18653/v1/2024.acl-long.747}.
\newblock URL \url{https://aclanthology.org/2024.acl-long.747/}.

\bibitem[Mao et~al.(2026)Mao, Liu, Liu, Tan, Shi, Wu, Zhang, and
  Wang]{mao2026comam}
Wenyu Mao, Haoyang Liu, Zhao Liu, Haosong Tan, Yaorui Shi, Jiancan Wu,
  An~Zhang, and Xiang Wang.
\newblock Collaborative multi-agent optimization for personalized memory
  system.
\newblock arXiv preprint arXiv:2603.12631v1, 2026.
\newblock URL \url{https://arxiv.org/abs/2603.12631v1}.

\bibitem[{memodb-io}(2024)]{memobase2024}
{memodb-io}.
\newblock {MemoBase}: User profile-based long-term memory for {AI} chatbot
  applications.
\newblock GitHub software repository, 2024.
\newblock URL \url{https://github.com/memodb-io/memobase}.
\newblock Accessed September 21, 2026.

\bibitem[Owolabi(2026)]{owolabi2026socialmembench}
Olukunle Owolabi.
\newblock {SocialMemBench}: Are {AI} memory systems ready for social group
  settings?
\newblock arXiv preprint arXiv:2605.17789, 2026.
\newblock URL \url{https://arxiv.org/abs/2605.17789}.

\bibitem[Packer et~al.(2023)Packer, Wooders, Lin, Fang, Patil, Stoica, and
  Gonzalez]{packer2023memgpt}
Charles Packer, Sarah Wooders, Kevin Lin, Vivian Fang, Shishir~G. Patil, Ion
  Stoica, and Joseph~E. Gonzalez.
\newblock {MemGPT}: Towards {LLMs} as operating systems.
\newblock arXiv preprint arXiv:2310.08560, 2023.
\newblock URL \url{https://arxiv.org/abs/2310.08560}.

\bibitem[Park et~al.(2023)Park, O'Brien, Cai, Morris, Liang, and
  Bernstein]{park2023generative}
Joon~Sung Park, Joseph~C. O'Brien, Carrie~J. Cai, Meredith~Ringel Morris, Percy
  Liang, and Michael~S. Bernstein.
\newblock Generative agents: Interactive simulacra of human behavior.
\newblock In \emph{Proceedings of the 36th Annual ACM Symposium on User
  Interface Software and Technology}, pp.\  10:1--10:22, San Francisco,
  California, USA, 2023. Association for Computing Machinery.
\newblock \doi{10.1145/3586183.3606763}.

\bibitem[Rasmussen et~al.(2025)Rasmussen, Paliychuk, Beauvais, Ryan, and
  Chalef]{rasmussen2025zep}
Preston Rasmussen, Pavlo Paliychuk, Travis Beauvais, Jack Ryan, and Daniel
  Chalef.
\newblock {Zep}: A temporal knowledge graph architecture for agent memory.
\newblock arXiv preprint arXiv:2501.13956, 2025.
\newblock URL \url{https://arxiv.org/abs/2501.13956}.

\bibitem[Robertson \& Zaragoza(2009)Robertson and Zaragoza]{robertson2009bm25}
Stephen Robertson and Hugo Zaragoza.
\newblock The probabilistic relevance framework: {BM25} and beyond.
\newblock \emph{Foundations and Trends in Information Retrieval}, 3\penalty0
  (4):\penalty0 333--389, 2009.
\newblock \doi{10.1561/1500000019}.

\bibitem[Sarthi et~al.(2024)Sarthi, Abdullah, Tuli, Khanna, Goldie, and
  Manning]{sarthi2024raptor}
Parth Sarthi, Salman Abdullah, Aditi Tuli, Shubh Khanna, Anna Goldie, and
  Christopher~D. Manning.
\newblock {RAPTOR}: Recursive abstractive processing for tree-organized
  retrieval.
\newblock In \emph{Proceedings of the International Conference on Learning
  Representations}, 2024.

\bibitem[Shao et~al.(2024)Shao, Wang, Zhu, Xu, Song, Bi, Zhang, Zhang, Li, Wu,
  and Guo]{shao2024deepseekmath}
Zhihong Shao, Peiyi Wang, Qihao Zhu, Runxin Xu, Junxiao Song, Xiao Bi, Haowei
  Zhang, Mingchuan Zhang, Y.~K. Li, Y.~Wu, and Daya Guo.
\newblock {DeepSeekMath}: Pushing the limits of mathematical reasoning in open
  language models.
\newblock arXiv preprint arXiv:2402.03300, 2024.
\newblock URL \url{https://arxiv.org/abs/2402.03300}.

\bibitem[Shinn et~al.(2023)Shinn, Cassano, Gopinath, Narasimhan, and
  Yao]{shinn2023reflexion}
Noah Shinn, Federico Cassano, Ashwin Gopinath, Karthik Narasimhan, and Shunyu
  Yao.
\newblock Reflexion: Language agents with verbal reinforcement learning.
\newblock In \emph{Advances in Neural Information Processing Systems 36}, New
  Orleans, Louisiana, USA, 2023. Neural Information Processing Systems
  Foundation.
\newblock URL
  \url{https://papers.nips.cc/paper_files/paper/2023/hash/1b44b878bb782e6954cd888628510e90-Abstract-Conference.html}.

\bibitem[Tan et~al.(2025)Tan, Zhang, Ma, Chen, Dai, and Dong]{tan2025membench}
Haoran Tan, Zeyu Zhang, Chen Ma, Xu~Chen, Quanyu Dai, and Zhenhua Dong.
\newblock {MemBench}: Towards more comprehensive evaluation on the memory of
  {LLM}-based agents.
\newblock In \emph{Findings of the Association for Computational Linguistics:
  ACL 2025}, pp.\  19336--19352, Vienna, Austria, 2025. Association for
  Computational Linguistics.
\newblock \doi{10.18653/v1/2025.findings-acl.989}.
\newblock URL \url{https://aclanthology.org/2025.findings-acl.989/}.

\bibitem[Tang et~al.(2026)Tang, Yu, Xiao, Wen, Li, Zhou, Wang, Wang, Huang,
  Deng, Sun, and Zhang]{tang2026mnemis}
Zihao Tang, Xin Yu, Ziyu Xiao, Zengxuan Wen, Zelin Li, Jiaxi Zhou, Hualei Wang,
  Haohua Wang, Haizhen Huang, Weiwei Deng, Feng Sun, and Qi~Zhang.
\newblock {Mnemis}: Dual-route retrieval on hierarchical graphs for long-term
  {LLM} memory.
\newblock In \emph{Proceedings of the 64th Annual Meeting of the Association
  for Computational Linguistics (ACL 2026)}, pp.\  23914--23928, San Diego,
  California, USA, 2026. Association for Computational Linguistics.
\newblock \doi{10.18653/v1/2026.acl-long.1096}.
\newblock URL \url{https://aclanthology.org/2026.acl-long.1096/}.

\bibitem[Wan \& Ma(2025)Wan and Ma]{wan2025storybench}
Luanbo Wan and Weizhi Ma.
\newblock Storybench: A dynamic benchmark for evaluating long-term memory with
  multi turns.
\newblock arXiv preprint arXiv:2506.13356, 2025.
\newblock URL \url{https://arxiv.org/abs/2506.13356}.

\bibitem[Wang et~al.(2026)Wang, Mao, Wu, Wang, and He]{wang2026r2mem}
Xinyuan Wang, Wenyu Mao, Junkang Wu, Xiang Wang, and Xiangnan He.
\newblock {R$^2$-Mem}: Reflective experience for memory search.
\newblock arXiv preprint arXiv:2605.13486, 2026.
\newblock URL \url{https://arxiv.org/abs/2605.13486}.

\bibitem[Wang \& Chen(2025)Wang and Chen]{wang2025mirix}
Yu~Wang and Xi~Chen.
\newblock {MIRIX}: Multi-agent memory system for {LLM}-based agents.
\newblock arXiv preprint arXiv:2507.07957, 2025.
\newblock URL \url{https://arxiv.org/abs/2507.07957}.

\bibitem[Wang et~al.(2024)Wang, Gao, Chen, Jiang, Li, Yang, Yin, Li, Li, Yin,
  Shang, and McAuley]{wang2024memoryllm}
Yu~Wang, Yifan Gao, Xiusi Chen, Haoming Jiang, Shiyang Li, Jingfeng Yang,
  Qingyu Yin, Zheng Li, Xian Li, Bing Yin, Jingbo Shang, and Julian McAuley.
\newblock {MemoryLLM}: Towards self-updatable large language models.
\newblock In \emph{Proceedings of the 41st International Conference on Machine
  Learning}, volume 235, pp.\  50453--50466, Vienna, Austria, 2024. PMLR.
\newblock URL \url{https://proceedings.mlr.press/v235/wang24s.html}.

\bibitem[Wang et~al.(2025{\natexlab{a}})Wang, Krotov, Hu, Gao, Zhou, McAuley,
  Gutfreund, Feris, and He]{wang2025mplus}
Yu~Wang, Dmitry Krotov, Yuanzhe Hu, Yifan Gao, Wangchunshu Zhou, Julian
  McAuley, Dan Gutfreund, Rogerio Feris, and Zexue He.
\newblock {M+}: Extending {MemoryLLM} with scalable long-term memory.
\newblock In \emph{Proceedings of the 42nd International Conference on Machine
  Learning}, volume 267, pp.\  63308--63323, Vancouver, Canada,
  2025{\natexlab{a}}. PMLR.
\newblock URL \url{https://proceedings.mlr.press/v267/wang25au.html}.

\bibitem[Wang et~al.(2025{\natexlab{b}})Wang, Takanobu, Liang, Mao, Hu,
  McAuley, and Wu]{wang2025memalpha}
Yu~Wang, Ryuichi Takanobu, Zhiqi Liang, Yuzhen Mao, Yuanzhe Hu, Julian McAuley,
  and Xiaojian Wu.
\newblock Mem-alpha: Learning memory construction via reinforcement learning.
\newblock arXiv preprint arXiv:2509.25911, 2025{\natexlab{b}}.
\newblock URL \url{https://arxiv.org/abs/2509.25911}.

\bibitem[Wu et~al.(2025)Wu, Wang, Yu, Zhang, Chang, and Yu]{wu2025longmemeval}
Di~Wu, Hongwei Wang, Wenhao Yu, Yuwei Zhang, Kai-Wei Chang, and Dong Yu.
\newblock {LongMemEval}: Benchmarking chat assistants on long-term interactive
  memory.
\newblock In \emph{Proceedings of the International Conference on Learning
  Representations}, 2025.
\newblock URL \url{https://openreview.net/forum?id=pZiyCaVuti}.

\bibitem[Xu et~al.(2026)Xu, Chen, Fang, Zhong, Yao, Zhu, Du, and
  Deng]{xu2026structmem}
Buqiang Xu, Yijun Chen, Jizhan Fang, Ruobin Zhong, Yunzhi Yao, Yuqi Zhu, Lun
  Du, and Shumin Deng.
\newblock {StructMem}: Structured memory for long-horizon behavior in {LLM}s.
\newblock In \emph{Proceedings of the 64th Annual Meeting of the Association
  for Computational Linguistics (Volume 2: Short Papers)}, pp.\  122--146, San
  Diego, California, USA, 2026. Association for Computational Linguistics.
\newblock \doi{10.18653/v1/2026.acl-short.12}.
\newblock URL \url{https://aclanthology.org/2026.acl-short.12/}.

\bibitem[Xu et~al.(2025)Xu, Liang, Mei, Gao, Tan, and Zhang]{xu2025amem}
Wujiang Xu, Zujie Liang, Kai Mei, Hang Gao, Juntao Tan, and Yongfeng Zhang.
\newblock {A-MEM}: Agentic memory for {LLM} agents.
\newblock In \emph{Advances in Neural Information Processing Systems 38}, pp.\
  20004--20031, San Diego, California, USA, 2025. Neural Information Processing
  Systems Foundation.
\newblock \doi{10.52202/085713-0593}.

\bibitem[Yan et~al.(2026{\natexlab{a}})Yan, Bahloul, Nie, Schwarzmann,
  Trivisonno, Tresp, and Ma]{yan2026memoryr2}
Sikuan Yan, Ahmed Bahloul, Ercong Nie, Susanna Schwarzmann, Riccardo
  Trivisonno, Volker Tresp, and Yunpu Ma.
\newblock Memory-{R2}: Fair credit assignment for long-horizon memory-augmented
  {LLM} agents.
\newblock arXiv preprint arXiv:2605.21768, 2026{\natexlab{a}}.
\newblock URL \url{https://arxiv.org/abs/2605.21768}.

\bibitem[Yan et~al.(2026{\natexlab{b}})Yan, Yang, Huang, Nie, Ding, Li, Ma, Bi,
  Kersting, Pan, Sch{\"u}tze, Tresp, and Ma]{yan2025memoryr1}
Sikuan Yan, Xiufeng Yang, Zuchao Huang, Ercong Nie, Zifeng Ding, Zonggen Li,
  Xiaowen Ma, Jinhe Bi, Kristian Kersting, Jeff~Z. Pan, Hinrich Sch{\"u}tze,
  Volker Tresp, and Yunpu Ma.
\newblock Memory-{R1}: Enhancing large language model agents to manage and
  utilize memories via reinforcement learning.
\newblock In \emph{Proceedings of the 64th Annual Meeting of the Association
  for Computational Linguistics (ACL 2026)}, pp.\  12805--12825, San Diego,
  California, USA, 2026{\natexlab{b}}. Association for Computational
  Linguistics.
\newblock \doi{10.18653/v1/2026.acl-long.583}.
\newblock URL \url{https://aclanthology.org/2026.acl-long.583/}.

\bibitem[Yang et~al.(2026)Yang, Lai, Wang, Chang, Harari, and
  Gabrilovich]{yang2026groupmembench}
Jingbo Yang, Kwei-Herng Lai, Xiaowen Wang, Shiyu Chang, Yaar Harari, and
  Evgeniy Gabrilovich.
\newblock {GroupMemBench}: Benchmarking {LLM} agent memory in multi-party
  conversations.
\newblock arXiv preprint arXiv:2605.14498, 2026.
\newblock URL \url{https://arxiv.org/abs/2605.14498}.

\bibitem[Yin \& Tang(2026)Yin and Tang]{yin2026defermem}
Jianing Yin and Tan Tang.
\newblock {DeferMem}: Query-time evidence distillation via reinforcement
  learning for long-term memory {QA}.
\newblock arXiv preprint arXiv:2605.22411, 2026.
\newblock URL \url{https://arxiv.org/abs/2605.22411}.

\bibitem[Yu et~al.(2026)Yu, Yao, Xie, Tan, Feng, Li, and Wu]{yu2026agentic}
Yi~Yu, Liuyi Yao, Yuexiang Xie, Qingquan Tan, Jiaqi Feng, Yaliang Li, and
  Libing Wu.
\newblock Agentic memory: Learning unified long-term and short-term memory
  management for large language model agents.
\newblock In \emph{Proceedings of the 64th Annual Meeting of the Association
  for Computational Linguistics (ACL 2026)}, pp.\  21457--21483, San Diego,
  California, USA, 2026. Association for Computational Linguistics.
\newblock \doi{10.18653/v1/2026.acl-long.981}.
\newblock URL \url{https://aclanthology.org/2026.acl-long.981/}.

\bibitem[Yue et~al.(2026)Yue, Hu, Sheng, Zhou, Zhang, Liu, Guo, and
  Deng]{yue2026hypermem}
Juwei Yue, Chuanrui Hu, Jiawei Sheng, Zuyi Zhou, Wenyuan Zhang, Tingwen Liu,
  Li~Guo, and Yafeng Deng.
\newblock {HyperMem}: Hypergraph memory for long-term conversations.
\newblock In \emph{Proceedings of the 64th Annual Meeting of the Association
  for Computational Linguistics (ACL 2026)}, pp.\  35237--35254, San Diego,
  California, USA, 2026. Association for Computational Linguistics.
\newblock \doi{10.18653/v1/2026.acl-long.1627}.
\newblock URL \url{https://aclanthology.org/2026.acl-long.1627/}.

\bibitem[Zhang et~al.(2025)Zhang, Fu, Wang, Wan, Yu, and Yan]{zhang2025gmemory}
Guibin Zhang, Muxin Fu, Kun Wang, Guancheng Wan, Miao Yu, and Shuicheng Yan.
\newblock {G-Memory}: Tracing hierarchical memory for multi-agent systems.
\newblock In \emph{Advances in Neural Information Processing Systems 38}, pp.\
  14587--14617, San Diego, California, USA, 2025. Neural Information Processing
  Systems Foundation.
\newblock \doi{10.52202/085713-0439}.

\bibitem[Zhang et~al.(2026)Zhang, Huang, Liu, Yang, Zhao, Wang, and
  Xie]{zhang2026deltamem}
Qi~Zhang, Shen Huang, Chu Liu, Shouqing Yang, Junbo Zhao, Haobo Wang, and
  Pengjun Xie.
\newblock {DeltaMem}: Towards agentic memory management via reinforcement
  learning.
\newblock arXiv preprint arXiv:2604.01560, 2026.
\newblock URL \url{https://arxiv.org/abs/2604.01560}.

\bibitem[Zhong et~al.(2024)Zhong, Guo, Gao, Ye, and Wang]{zhong2023memorybank}
Wanjun Zhong, Lianghong Guo, Qiqi Gao, He~Ye, and Yanlin Wang.
\newblock {MemoryBank}: Enhancing large language models with long-term memory.
\newblock \emph{Proceedings of the AAAI Conference on Artificial Intelligence},
  38\penalty0 (17):\penalty0 19724--19731, 2024.
\newblock \doi{10.1609/aaai.v38i17.29946}.

\end{thebibliography}
\bibliographystyle{speakermemr1}

\clearpage
\appendix
\onecolumn
\section*{Appendix}

This appendix provides the data structures and algorithms, implementation and evaluation protocols, R1 training details, reproduction checklist, complete category-level results, LoCoMo breakdown, retrieval-budget sensitivity, and all key prompts.

\section{Implementation Details}
\label{app:implementation}

The system first deterministically stores verbatim messages and the channel roster, after which the Writer produces derived records with source provenance. System~1 selects verbatim evidence and preserves speaker attribution; System~2 retrieves structured records by per-speaker or group rows and follows provenance pointers back to the source text. Updates append new nodes and mark old nodes as superseded rather than overwriting history.

\paragraph{Fixed query budgets.}
All main experiments, ablations, and the controlled SFT/R1 Writer evaluation use the same query configuration: System~1 has final top-$k=10$ with the default recall pool recall-$n=40$, and System~2 takes $k=2$ records per owner row with source-$k=1$. Select ranks the recalled System~1 candidates, after which Expand may add neighboring messages; when ASK is enabled, it may add a second recalled pool before the final reranking and top-$k$ truncation. The two paths use independent budgets.

\paragraph{Execution order.}
Project parses the target roster, issue/event, head/full temporal mode, and relation constraints from the question; Select forms candidates for System~1 and System~2 separately. Anchor then retains source, owner, event, time, and provenance; Separate expands PERSON/GROUP rows from the roster; Resolve distinguishes issues, parallel events, and temporal versions within rows; and Compose organizes System~1 and System~2 evidence. Empty derived rows fall back to the corresponding person's verbatim messages, and explicit empty rows are retained when no evidence exists. The answerer receives only the final evidence set, while R1 training updates only the Writer. Complete prompts and reward decomposition are given in Appendices~\ref{app:prompts} and~\ref{app:additional}.

\subsection{Five-layer memory schema}
\label{app:five-layer-memory}

The following table gives the detailed schema behind the concise description in Section~\ref{sec:five-layer-memory}. It records the implementation layer names, scope, information retained, and the role each layer plays during query-time evidence construction. System~2 records carry source/owner fields and provenance pointers (implemented as \texttt{from\_ids}) to System~1 messages.

\begin{table}[!t]
\caption{Detailed five-layer schema of \method{}. System~1 is the verbatim track; System~2 contains four derived layers. The code names match the Writer contract and stored records.}
\label{tab:five-layers}
\centering
\small
\renewcommand{\arraystretch}{1.08}
\setlength{\tabcolsep}{3pt}
\begin{tabularx}{\textwidth}{@{}>{\raggedright\arraybackslash}p{1.35cm}>{\centering\arraybackslash}p{1.3cm}>{\raggedright\arraybackslash}p{3.3cm}>{\raggedright\arraybackslash}X>{\raggedright\arraybackslash}p{3.15cm}@{}}
\toprule
\textbf{Track} & \textbf{Scope} & \textbf{Layer (code name)} & \textbf{Stored information} & \textbf{Query role and provenance} \\
\midrule
System~1 & PERSON & \makecell[l]{Episodic\\\texttt{per\_speaker\_}\\\texttt{episodic}} & Verbatim message text, speaker, session/turn, timestamp, channel, and message identifier; preserves wording and local context. & Supplies exact wording and neighboring context to System~1 retrieval; messages are deterministically appended and serve as the provenance target. \\
\midrule
System~2 & PERSON & \makecell[l]{Core\\\texttt{per\_speaker\_}\\\texttt{core}} & Stable identity, facts, stances, and recurring behavior; summarizes a person's persistent state. & Owner-row retrieval for personal-state questions; each record keeps source/owner and points to supporting System~1 messages through \texttt{from\_ids}. \\
System~2 & PERSON & \makecell[l]{Profile\\\texttt{per\_speaker\_}\\\texttt{profile}} & Observations about a person and cross-person beliefs or perceptions. & Owner-row retrieval for relational questions; separate source (who provides the information) from owner (whom it concerns), with \texttt{from\_ids} provenance. \\
System~2 & GROUP & \makecell[l]{Interaction\\\texttt{group\_}\\\texttt{interaction}} & Cross-speaker events, relations, and group decisions with their outcomes. & Group-row retrieval for multi-person events and decisions; GROUP scope and source/owner are retained with provenance links. \\
System~2 & GROUP & \makecell[l]{Insight\\\texttt{group\_}\\\texttt{insight}} & Group norms, consensus, and exceptions that distinguish collective from individual behavior. & Group-row retrieval for norms and shared beliefs; records remain linked to the supporting System~1 messages. \\
\bottomrule
\end{tabularx}
\end{table}

For all four System~2 layers, an update appends a new node and links it to the prior state rather than overwriting the source record; the head/full query mode selects the current state or the complete update chain (Appendix~\ref{app:algorithms}).

\section{Data Structures and Algorithmic Details}
\label{app:algorithms}

This appendix supplements the algorithms, data structures, implementation parameters, complete prompts, and examples needed to reproduce the method but omitted from the main paper. This section gives the data structures and algorithmic flow; training parameters are in Appendix~\ref{app:additional}, the evaluation protocol is in Appendix~\ref{app:benchmarks}, and category-level results are in Appendix~\ref{app:multibench-detailed}.

\subsection{Memory Record Structure}

The following table lists persistent fields shared by the verbatim and derived tracks.

\begin{table}[!t]
\caption{Persistent record fields in \method{}. The verbatim and derived tracks share this schema; the Writer generates only action fields, while the system fills the remaining coordinates from messages and the current state.}
\label{tab:memory-schema}
\centering\scriptsize
\setlength{\tabcolsep}{3pt}
\begin{tabular}{>{\raggedright\arraybackslash}p{2.35cm}>{\raggedright\arraybackslash}p{2.45cm}>{\raggedright\arraybackslash}p{8.05cm}}
\toprule
\textbf{Field} & \textbf{Type/value} & \textbf{Meaning} \\
\midrule
\texttt{entry\_id} & string & Unique record identifier; also the target of UPDATE, chain, and source-provenance pointers. \\
\texttt{content} & string & Original message or a single derived fact, stance, observation, decision, or relation. \\
\texttt{owner}, \texttt{source} & PERSON/GROUP, speaker & \texttt{owner} is whom the record concerns; \texttt{source} is who stated or observed the information. \\
\texttt{layer} & Five discrete values & verbatim layer \texttt{per\_speaker\_episodic}; personal derived layers \texttt{core/profile}; group layers \texttt{interaction/insight}. \\
\texttt{utype} & Six discrete values & \texttt{utterance}, \texttt{fact}, \texttt{stance}, \texttt{observation}, \texttt{decision}, \texttt{relation}. \\
\texttt{session}, \texttt{turn} & string, integer & Channel/session and within-session turn, used for local verbatim expansion. \\
\texttt{turn\_created}, \texttt{ts} & integer, timestamp & Global write order and real-time coordinate. \\
\texttt{from\_ids} & list[string] & Verbatim records supporting a derived record, enabling provenance back to the source text. \\
\texttt{links}, \texttt{superseded\_by} & list/string & Old-to-new chains for non-destructive UPDATE and other relations. \\
\texttt{confidence} & $[0,1]$ & Writing confidence and retrieval-weighting signal. \\
\bottomrule
\end{tabular}
\end{table}

\subsection{Online Writing}

For an arriving message segment $X_t$, the system executes the following steps:

\begin{enumerate}[leftmargin=*,itemsep=1pt]
  \item Write every message verbatim into \texttt{per\_speaker\_episodic}, recording the speaker, session, turn, timestamp, and message identifier; update the channel roster.
  \item Read the current heads of the four derived layers and form the Writer input together with $X_t$.
  \item The Writer generates at most the allowed number of \textsc{Add}/\textsc{Update}/\textsc{Noop} actions using the format and operation requirements in Appendix~\ref{app:prompts}.
  \item Validate JSON, action type, source/owner, layer, entry ID, and within-batch duplicates. For \textsc{Add}, create a new record and attach supporting-message \texttt{from\_ids}; for \textsc{Update}, create a new node, inherit the original coordinates, and connect it to the old state using \texttt{links/superseded\_by}. Old nodes are not deleted.
  \item Write valid nodes to persistent storage and update the vector index; \textsc{Noop} leaves the derived track unchanged.
\end{enumerate}

This process separates ``what to write'' from ``coordinates completed by the system'': the Writer decides what to record, while deterministic code fills message coordinates, provenance, and update chains.

\paragraph{Action state transitions.} The Writer input consists of the current message segment $X_t$, the roster, and the current heads of the four derived layers; an \textsc{Update} action must additionally carry the \texttt{entry\_id} of the node to update. The system validates that this ID resolves to an existing derived record and reuses its owner, source, and layer coordinates, then writes the new content and time and connects old and new states through \texttt{links}/\texttt{superseded\_by}. If the ID is invalid, Update is rejected and treated as Add. The old node remains stored; \texttt{head} versus \texttt{full} controls whether querying returns only the current node or the complete chain. Noop creates no derived node.

\subsection{Dual-Track Composition at Query Time}

Given a question $q$ and channel roster, the query proceeds as follows:

\begin{enumerate}[leftmargin=*,itemsep=1pt]
  \item \textbf{Project:} output the issue, PERSON/GROUP rows, and \texttt{head/full}.
  \item \textbf{System~1:} retrieve a default pool of $n=40$ candidates from the verbatim track; Select ranks the pool for necessity, and Expand fetches neighboring source messages along selected hits' session/turn. If final evidence remains insufficient and ASK is enabled, Sufficiency generates one query term, another candidate pool is recalled and merged, and the combined pool is reranked. Only after these operations is the final set truncated to top-$k$.
  \item \textbf{System~2:} expand rows into concrete roster rows. Retrieve derived records by issue within each row and collapse or expand update chains according to \texttt{head/full}; source supplements do not consume the owner-row budget. Empty derived rows can fall back to the row's verbatim messages, and explicit empty rows are retained when evidence is unavailable.
  \item \textbf{Compose:} preserve the owner, source, time, and provenance of every record, and pass System~1 verbatim evidence alongside System~2 structured evidence to the frozen answerer.
\end{enumerate}

The two retrieval paths use independent budgets, and System~2 does not participate in System~1 pre-pruning. Structured memory therefore determines which objects, scopes, and states must be checked, while verbatim memory supplies exact wording and details.

\section{Benchmarks and Evaluation Protocol}
\label{app:benchmarks}

The table summarizes the number of questions, metrics and location of the category-level results reported in this paper.

\begin{table}[!t]
\caption{Evaluation scope, number of questions, primary metric, and reporting location.}
\label{tab:benchmark-scope}
\centering\small
\begin{tabularx}{\textwidth}{@{}l r >{\raggedright\arraybackslash}p{3.2cm} >{\raggedright\arraybackslash}X@{}}
\toprule
\textbf{Benchmark} & \textbf{Questions} & \textbf{Metrics} & \textbf{Reporting scope} \\
\midrule
GroupMemBench & 745 & Question-level binary accuracy & Six-category accuracy and token-F1; the GPT-5.6-luna setting is reported separately \\
SocialMemBench & 1,031 & Question-level binary accuracy & Q1--Q9; MeanQ/MeanN and Q1--Q9 category results are also reported \\
EverMemBench & 2,400 & Question-level binary accuracy & EverMind-AI nine-behavior accuracy and weighted overall result \\
LoCoMo & 1,986 & Question-level binary accuracy & Five-category accuracy and the published comparison protocol excluding AD \\
\bottomrule
\end{tabularx}
\end{table}

All main \method{} experiments and ablations use the default setting: System~1 top-$k=10$ and System~2 $k=2$/source-$k=1$. Full context is included only as a reference upper bound on SocialMem. Both DeepSeek-V4-Flash and GPT-5.6-luna rows use GPT-4o-mini for question-level judging; the EverMemBench cross-configuration results and judge settings are given in Table~\ref{tab:evermem-cross-config}. token-F1 is computed directly from the answer and gold answer and does not depend on a judge; the two result types are reported separately. For the primary three-benchmark comparison, each system follows its official code, recommended configuration, and official prompt; we standardize only the data, metric definitions, GPT-4o-mini judge, and evaluation interface, while speaker/source metadata may be retained where supported. Detailed procedures and result tables are provided in the supplementary material.

\paragraph{Acc. and token-F1.}
Let $Q$ be the number of evaluated questions and $j_i\in\{0,1\}$ the judgment for question $i$: $j_i=1$ only when the answer is judged correct, and $0$ otherwise. Question-level binary accuracy is
\begin{equation}
\mathrm{Acc.}=\frac{100}{Q}\sum_{i=1}^{Q}j_i.
\label{eq:binary-accuracy}
\end{equation}
token-F1 is computed per question and then averaged with equal question weights. Following the supplementary implementation, predictions and references are lowercased, stripped of surrounding whitespace and trailing periods, normalized for repeated whitespace, and split on whitespace; duplicate tokens are removed to form sets $A_i$ and $G_i$, respectively. Define
\begin{equation}
p_i=\frac{|A_i\cap G_i|}{|A_i|},\quad
r_i=\frac{|A_i\cap G_i|}{|G_i|},\quad
f_i=\frac{2p_ir_i}{p_i+r_i},\quad
\mathrm{token\text{-}F1}=\frac{100}{Q}\sum_{i=1}^{Q}f_i.
\label{eq:token-f1}
\end{equation}
If either set is empty or their intersection is empty, we directly set $f_i=0$ to avoid division by zero. This implementation uses token sets rather than multisets retaining repeated occurrences; it measures lexical overlap rather than semantic correctness. Both Acc. and token-F1 are reported as percentages; category results use the same definitions averaged over questions in that category.

\paragraph{MeanQ and MeanN.}
SocialMemBench first assigns each question a score $s_q\in[0,1]$: the 214 multiple-choice questions receive 0/1 for exact option matching, while the 817 open-ended questions receive partial credit under the benchmark's itemized scoring protocol; per-person stance questions are scored by the fraction of members whose stances are recovered correctly~\citep{owolabi2026socialmembench}. Let $\mathcal{Q}_n$ denote the question set for network $n$, with $N=43$ networks. Then
\begin{equation}
\operatorname{MeanQ}=\frac{1}{\sum_n|\mathcal{Q}_n|}\sum_{n=1}^{N}\sum_{q\in\mathcal{Q}_n}s_q,
\qquad
\operatorname{MeanN}=\frac{1}{N}\sum_{n=1}^{N}\frac{1}{|\mathcal{Q}_n|}\sum_{q\in\mathcal{Q}_n}s_q.
\label{eq:socialmem-means}
\end{equation}
MeanQ gives every question equal weight, so networks with more questions contribute more; MeanN first computes a mean for each network and then weights all networks equally, reducing the effect of unequal network sizes. Both are means of item-level partial scores, not the proportion of questions answered completely correctly, and neither is interchangeable with question-level binary accuracy. Our 95\% CI applies only to MeanN: the 43 network means are the bootstrap units, and we use 10,000 resamples and take the 2.5th and 97.5th percentiles of the resulting distribution.

\section{Reproduction Checklist and Experimental Boundaries}
\label{app:reproducibility}

This section summarizes the experimental boundaries using common reproducibility checks. Detailed algorithms, hyperparameters, and prompts are given in Appendices~\ref{app:algorithms}, \ref{app:additional}, and \ref{app:prompts}. The table distinguishes the language models used across system stages from question-level judging settings.

\begin{table}[!t]
\caption{Reproduction checklist for \method{}.}
\label{tab:reproducibility-checklist}
\centering
\scriptsize
\setlength{\tabcolsep}{2.5pt}
\begin{tabular}{>{\raggedright\arraybackslash}p{2.3cm}>{\raggedright\arraybackslash}p{4.0cm}>{\raggedright\arraybackslash}p{7.0cm}}
\toprule
\textbf{Item} & \textbf{Fixed setting} & \textbf{Reproduction location and notes} \\
\midrule
Data and splits & GroupMemBench 745; SocialMemBench 1,031; EverMemBench 2,400; LoCoMo 1,986 & Benchmark names, question counts, categories, and comparison conventions are given in Appendix~\ref{app:benchmarks}; R1 training and held-out networks are separated at the network level. \\
Model roles & DeepSeek-V4-Flash or GPT-5.6-luna is used across memory construction, retrieval, and answer generation & The main results in Table~\ref{tab:main-results} and the complete category tables identify the model configuration; both DeepSeek-V4-Flash and GPT-5.6-luna rows use GPT-4o-mini for question-level judging; the EverMemBench cross-configuration results identify the judge in Table~\ref{tab:evermem-cross-config}.\\
Memory writing & System~1 deterministically appends verbatim messages; System~2 uses the Qwen2.5-3B Writer to generate four types of derived records & The five-layer schema, action-format requirements, and non-destructive update chains are given in Appendix~\ref{app:algorithms}. \\
Query budgets & System~1: recall-$n=40$, final top-$k=10$; Expand at most twice; ASK at most once with 20 retrieved items; System~2: $k=2$/source-$k=1$ & Complete input/output specifications for Project, Select/Expand, Sufficiency/ASK, and within-row selection are given in Appendix~\ref{app:prompts}. \\
R1 training & $G=8$ trajectories; 30 rollout rounds; 2 epochs per update; learning rate $10^{-6}$; temperature .8; top-$p=.95$ & SpeakerLevenshtein (speaker-conditioned matching), speaker-conditioned GRPO, reward weights, KL, and data scale are given in Appendix~\ref{app:additional}. \\
Evaluation convention & We report question-level binary accuracy and token-F1; SocialMem additionally reports MeanQ/MeanN & Both DeepSeek-V4-Flash and GPT-5.6-luna rows use GPT-4o-mini for judging; EverMemBench cross-configuration results and judge settings are given in Table~\ref{tab:evermem-cross-config}; token-F1 is computed directly from answers and gold answers without a judge. The two result types are reported separately.\\
\bottomrule
\end{tabular}
\end{table}

\paragraph{Primary evaluation protocol.} For the three multi-party benchmark comparisons, each system follows its official code, recommended configuration, and official prompt. We standardize only the data, metric definitions, judge, and evaluation interface; implementations may preserve speaker/source metadata where supported. Detailed execution procedures and result tables are provided in the supplementary material.

\paragraph{Implementation boundaries.}
Main benchmark results and the controlled SFT/R1 Writer evaluation use the same query configuration: System~1 top-$k=10$ with the default recall-$n=40$, System~2 $k=2$ per owner plus source-$k=1$. SocialMemBench MeanN uses 10,000 bootstrap resamples over 43 networks and reports a 95\% confidence interval. Because the R1 training set contains model-generated teacher trajectories and manually authored boundary networks, the appendix reports their sources, action statistics, and train/evaluation isolation. Datasets are used under the licenses and access terms of their original publishers; we report only experimental settings, data composition, and results.

\paragraph{Token accounting for final result artifacts.}
We count tokens from the prompts, messages, memories, and answers processed at each stage. Total tokens include ingestion input, ingestion output, QA input, and final answer output, and exclude the judge stage; average tokens per question use the corresponding benchmark size as denominator. The three benchmarks contain 4,176 questions: 1,031 SocialMem, 745 GroupMem, and 2,400 EverMem. The SpeakerMem Writer can emit multiple update actions in one call, enabling small-batch ingestion; we therefore use the deployed configuration, with session-level ingestion for SocialMem and Window25 for GroupMem/EverMem. For Mem0, A-MEM, and HippoRAG, larger input chunks can reduce memory quality, so we use chunk5: each ingestion call processes five messages before the system performs memory merging, conflict resolution, and subsequent steps. This retains limited batching while avoiding the very high token cost of fully serial ingestion, and provides the cost-comparison setting reported here.

\paragraph{Three-benchmark overview.} This table aggregates all processing stages. SpeakerMem uses session-level ingestion for SocialMem and Window25 for GroupMem/EverMem; the other memory systems ingest messages in chunk5 units.
\begin{table}[!t]
\caption{Token totals across the final result artifacts.}
\label{tab:token-accounting}
\centering
\scriptsize
\setlength{\tabcolsep}{2.5pt}
\resizebox{\linewidth}{!}{%
\begin{tabular}{lrrrrrrr}
\toprule
\textbf{System} & \textbf{Calls/units} & \textbf{Ingest in.} & \textbf{Ingest out.} & \textbf{QA input} & \textbf{Answer out.} & \textbf{Total} & \textbf{Avg./question} \\
\midrule
\rowcolor{gray!10}
SpeakerMem-R1+ASK & 7,212 & 25,546,058 & 721,200 & 13,865,187 & 430,530 & 40,562,975 & 9,713.36 \\\\
BM25 & 0 & 0 & 0 & 5,298,971 & 85,011 & 5,383,982 & 1,289.27 \\\\
Embed & 0 (local) & 0 & 0 & 5,096,215 & 96,764 & 5,192,979 & 1,243.53 \\\\
Mem0-chunk5 & 35,676 & 29,970,000 & 29,970,000 & 2,316,690 & 91,476 & 62,348,166 & 14,930.12 \\\\
A-MEM-chunk5 & 35,676 & 39,929,508 & 20,100,569 & 12,482,895 & 159,830 & 72,672,802 & 17,402.49 \\\\
HippoRAG-chunk5 & 35,676 & 30,400,001 & 30,400,001 & 11,506,209 & 155,857 & 72,462,068 & 17,352.03 \\\\
\bottomrule
\end{tabular}%
}
\end{table}
\paragraph{SocialMemBench breakdown.} SocialMemBench contains 1,031 questions. SpeakerMem ingests 348 sessions; the other systems ingest 1,471 message units under chunk5.
\begin{table}[!t]
\caption{SocialMemBench (1,031 questions)}
\label{tab:token-socialmem}
\centering
\scriptsize
\setlength{\tabcolsep}{2.5pt}
\resizebox{\linewidth}{!}{%
\begin{tabular}{lrrrrrrr}
\toprule
\textbf{System} & \textbf{Calls/units} & \textbf{Ingest in.} & \textbf{Ingest out.} & \textbf{QA input} & \textbf{Answer out.} & \textbf{Total} & \textbf{Avg./question} \\
\midrule
\rowcolor{gray!10}
SpeakerMem-R1+ASK & 348 & 845,605 & 34,800 & 846,404 & 75,247 & 1,802,056 & 1747.87 \\\\
BM25 & 0 & 0 & 0 & 1,308,247 & 20,988 & 1,329,235 & 1289.27 \\\\
Embed & 0 & 0 & 0 & 1,258,189 & 23,890 & 1,282,079 & 1243.53 \\\\
Mem0-chunk5 & 1,471 & 1,236,000 & 1,236,000 & 306,976 & 31,561 & 2,810,537 & 2726.03 \\\\
A-MEM-chunk5 & 1,471 & 1,646,400 & 828,800 & 1,328,327 & 73,915 & 3,877,442 & 3760.86 \\\\
HippoRAG-chunk5 & 1,471 & 1,253,473 & 1,253,473 & 1,032,295 & 71,444 & 3,610,685 & 3502.12 \\\\
\bottomrule
\end{tabular}%
}
\end{table}
\paragraph{GroupMemBench breakdown.} GroupMemBench contains 745 questions and 120,000 messages. SpeakerMem makes 4,815 Window25 Writer calls; the other systems ingest 24,000 chunk5 units.
\begin{table}[!t]
\caption{GroupMemBench (745 questions)}
\label{tab:token-groupmem}
\centering
\scriptsize
\setlength{\tabcolsep}{2.5pt}
\resizebox{\linewidth}{!}{%
\begin{tabular}{lrrrrrrr}
\toprule
\textbf{System} & \textbf{Calls/units} & \textbf{Ingest in.} & \textbf{Ingest out.} & \textbf{QA input} & \textbf{Answer out.} & \textbf{Total} & \textbf{Avg./question} \\
\midrule
\rowcolor{gray!10}
SpeakerMem-R1+ASK & 4,815 & 17,445,456 & 481,500 & 2,152,506 & 29,344 & 20,108,806 & 26991.69 \\\\
BM25 & 0 & 0 & 0 & 945338 & 15166 & 960504 & 1289.27 \\\\
Embed & 0 & 0 & 0 & 909167 & 17263 & 926430 & 1243.53 \\\\
Mem0-chunk5 & 24,000 & 20,160,000 & 20,160,000 & 253,800 & 20,582 & 40,594,382 & 54489.10 \\\\
A-MEM-chunk5 & 24,000 & 26,861,726 & 13,522,230 & 2,444,228 & 33,277 & 42,861,461 & 57532.16 \\\\
HippoRAG-chunk5 & 24,000 & 20,450,953 & 20,450,953 & 2,448,714 & 32,569 & 43,383,189 & 58232.47 \\\\
\bottomrule
\end{tabular}%
}
\end{table}
\paragraph{EverMemBench breakdown.} EverMemBench contains 2,400 questions and 51,023 messages. SpeakerMem makes 2,049 Window25 Writer calls; the other systems ingest 10,205 chunk5 units.
\begin{table}[!t]
\caption{EverMemBench (2,400 questions)}
\label{tab:token-evermem}
\centering
\scriptsize
\setlength{\tabcolsep}{2.5pt}
\resizebox{\linewidth}{!}{%
\begin{tabular}{lrrrrrrr}
\toprule
\textbf{System} & \textbf{Calls/units} & \textbf{Ingest in.} & \textbf{Ingest out.} & \textbf{QA input} & \textbf{Answer out.} & \textbf{Total} & \textbf{Avg./question} \\
\midrule
\rowcolor{gray!10}
SpeakerMem-R1+ASK & 2,049 & 7,254,997 & 204,900 & 10,866,277 & 325,939 & 18,652,113 & 7771.71 \\\\
BM25 & 0 & 0 & 0 & 3045386 & 48857 & 3094243 & 1289.27 \\\\
Embed & 0 & 0 & 0 & 2928859 & 55611 & 2984470 & 1243.53 \\\\
Mem0-chunk5 & 10,205 & 8,574,000 & 8,574,000 & 1,755,914 & 39,333 & 18,943,247 & 7893.02 \\\\
A-MEM-chunk5 & 10,205 & 11,421,382 & 5,749,539 & 8,710,340 & 52,638 & 25,933,899 & 10805.79 \\\\
HippoRAG-chunk5 & 10,205 & 8,695,575 & 8,695,575 & 8,025,200 & 51,844 & 25,468,194 & 10611.75 \\\\
\bottomrule
\end{tabular}%
}
\end{table}

\section{Additional R1 Training Details}
\label{app:additional}

\paragraph{Record matching with SpeakerLevenshtein (speaker-conditioned matching).}
Let the predicted and reference derived states be $M$ and $M^\star$, partitioned into buckets $M_p$ and $M_p^\star$ by owner, including a separate \texttt{GROUP} owner. The soft content match and coordinate gate between predicted record $i$ and reference record $j$ are
\begin{align}
\bar\kappa_{ij}
  &=\alpha_{\mathrm{tok}}F^{\mathrm{tok}}_1(c_i,c_j)
    +\alpha_{\mathrm{seq}}S_{\mathrm{seq}}(c_i,c_j),\\
\kappa_{ij}
  &=\mathbf{1}[o_i=o_j,\,s_i=s_j,\,\ell_i=\ell_j]
    \mathbf{1}[\bar\kappa_{ij}\ge0.50]\bar\kappa_{ij},
\label{eq:speaker-pair-score}
\end{align}
We use $\alpha_{\mathrm{tok}}=0.60$ and $\alpha_{\mathrm{seq}}=0.40$. For normalized token sequences $T_i$ and $T_j$, let $m_{ij}$ be the total number of matched tokens returned by \texttt{SequenceMatcher}; its ratio is
\begin{equation}
S_{\mathrm{seq}}(c_i,c_j)=\frac{2m_{ij}}{|T_i|+|T_j|},\qquad T_i=T(c_i),\;T_j=T(c_j).
\end{equation}
where $c,o,s,\ell$ denote content, owner, source, and memory layer, respectively; $S_{\mathrm{seq}}$ is the ratio returned by Python \texttt{SequenceMatcher.ratio()} on normalized token sequences: content is lowercased, \texttt{[a-z0-9]+} spans are tokenized, and each CJK character is a token; the tokens are joined with spaces before computing the ratio. The name follows the experimental implementation, but the score itself does not use standard Levenshtein edit distance. Semantically similar records with inconsistent coordinates receive no match score. Within each owner bucket, we perform maximum-weight one-to-one Hungarian alignment:
\begin{equation}
\mathcal{A}_p^\star=\arg\max_{\mathcal{A}\in\mathfrak{M}(M_p,M_p^\star)}
  \sum_{(i,j)\in\mathcal{A}}\kappa_{ij},\qquad
F_p=\frac{2\sum_{(i,j)\in\mathcal{A}_p^\star}\kappa_{ij}}
{|M_p|+|M_p^\star|},
\label{eq:speaker-hungarian}
\end{equation}
where $\mathfrak{M}$ is the set of one-to-one matchings. A prediction cannot match multiple reference facts, so $F_p$ penalizes both omissions and over-writing.

\paragraph{Update-chain reward.}
Correct state nodes do not imply correct update relations. For predicted and reference UPDATE edges, we first require matching owner and layer coordinates, then compare the old and new states separately:
\begin{equation}
\kappa^{\mathrm{chain}}_{ij}
=w_{\mathrm{old}}\bar\kappa(c_i^{\mathrm{old}},c_j^{\mathrm{old}})
+w_{\mathrm{new}}\bar\kappa(c_i^{\mathrm{new}},c_j^{\mathrm{new}}).
\end{equation}
We use $w_{\mathrm{old}}=0.50$ and $w_{\mathrm{new}}=0.50$. Applying the same one-to-one soft-F1 alignment to chain edges yields $R^{\mathrm{chain}}_{g,t}$, and the memory reward is
\begin{equation}
R^{\mathrm{mem}}_{g,t}
=w_{\mathrm{SL}}\Phi_{\mathrm{SL}}(M_{g,t},M_t^\star)
+w_{\mathrm{chain}}R^{\mathrm{chain}}_{g,t}.
\label{eq:memory-reward-full}
\end{equation}
We use $w_{\mathrm{SL}}=0.80$ and $w_{\mathrm{chain}}=0.20$.

\paragraph{Action validity and structural penalties.}
For action $t$ on candidate trajectory $g$,
\begin{equation}
R^{\mathrm{valid}}_{g,t}
=w_{\mathrm{json}}R^{\mathrm{json}}_{g,t}+w_{\mathrm{schema}}R^{\mathrm{schema}}_{g,t}
+w_{\mathrm{dup}}R^{\mathrm{dup}}_{g,t}+w_{\mathrm{stop}}R^{\mathrm{stop}}_{g,t}.
\label{eq:validity-reward-full}
\end{equation}
We use $w_{\mathrm{json}}=0.25$, $w_{\mathrm{schema}}=0.35$, $w_{\mathrm{dup}}=0.25$, and $w_{\mathrm{stop}}=0.15$. Here a candidate action list is denoted by a and n is its attempted-action count (from the execution report, or from the list length when omitted).
The four components \(R^{\mathrm{json}}_{g,t}\), \(R^{\mathrm{schema}}_{g,t}\), \(R^{\mathrm{dup}}_{g,t}\), and \(R^{\mathrm{stop}}_{g,t}\) are indexed by the trajectory and writing position (g,t). The JSON component is one iff the report marks the output as valid JSON.
The schema component is the clipped fraction of schema-valid actions among attempts; if the count is absent, it is inferred from attempts minus validation errors, and is zero for invalid JSON.
The duplicate component is the clipped fraction of unique action keys among attempts. ADD keys contain action, owner, source, layer, and normalized content; UPDATE keys contain action, entry ID, and normalized content; a NOOP key is its action type.
If the attempted-action count is zero, all four components and the final validity score are zero.

The stop component is one iff the output stops immediately after its last novel action, and is zero otherwise. A single NOOP is novel by definition. An ADD is novel when its key is neither repeated in the output nor already present in the pre-state. An UPDATE is novel only when its entry ID resolves to an active pre-state record and its normalized content differs from that record. The Writer action budget is
\begin{equation}
  n_{g,t}\leq\mathrm{max\_actions}(X_t)=\max(8,2|X_t|),
\end{equation}
where $|X_t|$ is the number of messages in the input segment; exceeding this
bound is reported as \texttt{too\_many\_actions} and contributes the count
term in the structural penalty below.
Let $z^{\mathrm{json}},z^{\mathrm{act}},z^{\mathrm{count}},z^{\mathrm{len}}\in\{0,1\}$ indicate invalid JSON, invalid actions, too many actions, and output reaching the length limit, respectively. Then
\begin{equation}
P_{g,t}=-\eta_{\mathrm{json}}z^{\mathrm{json}}_{g,t}-\eta_{\mathrm{act}}z^{\mathrm{act}}_{g,t}
-\eta_{\mathrm{count}}z^{\mathrm{count}}_{g,t}-\eta_{\mathrm{len}}z^{\mathrm{len}}_{g,t}.
\label{eq:structural-penalty}
\end{equation}
We use $\eta_{\mathrm{json}}=0.25$, $\eta_{\mathrm{act}}=0.15$, $\eta_{\mathrm{count}}=0.10$, and $\eta_{\mathrm{len}}=0.10$. Together these terms provide bounded dense structural signals for each transition.

\paragraph{Terminal QA gain.}
For each training network, we sort by question ID and uniformly sample at most four questions to form a fixed subset $\mathcal Q$, with binary judge $J(\hat a,a^\star)$. Define
\begin{equation}
\operatorname{QA}(E;M)=\frac{1}{|\mathcal Q|}\sum_{q\in\mathcal Q}
J\!\left(\operatorname{Ans}(q,E;M),a_q^\star\right).
\end{equation}
System~1+System~2 and System~1-only share the same System~1 evidence, so Eq.~\ref{eq:terminal-qa-delta} isolates the gain from the derived track. This terminal difference is propagated to every action on the trajectory with $\gamma^{T-1-t}$ and combined with local terms in Eq.~\ref{eq:sc-grpo-return}.

\paragraph{Transition-aligned group-relative advantages.}
We sample $G=8$ independent on-policy trajectories for the same network. The within-group mean and standard deviation at the same position $t$ are
\begin{equation}
\mu_t=\frac{1}{G}\sum_{g=1}^{G}r_{g,t},\qquad
\sigma_t=\sqrt{\frac{1}{G}\sum_{g=1}^{G}(r_{g,t}-\mu_t)^2}.
\end{equation}
The advantage is defined as
\begin{equation}
\hat A_{g,t}=
\begin{cases}
0, & \sigma_t<0.02,\\
\operatorname{clip}\!\left((r_{g,t}-\mu_t)/\sigma_t,-3,3\right),&\text{otherwise}.
\end{cases}
\label{eq:sc-grpo-advantage}
\end{equation}
Position-wise comparison prevents semantically different actions from trajectories of different lengths from sharing one normalization group. Zero-variance groups provide no effective relative signal and are therefore excluded from updates.

\paragraph{Clipped GRPO and KL.}
For completion token $y_{g,t,k}$, the importance ratio is
\begin{equation}
\rho_{g,t,k}(\theta)=
\exp\!\left(\log\pi_\theta(y_{g,t,k}\mid x_{g,t},y_{g,t,<k})
-\log\pi_{\mathrm{old}}(y_{g,t,k}\mid x_{g,t},y_{g,t,<k})\right).
\end{equation}
Equation~\ref{eq:sc-grpo-return} first gives the return for each transition, after which group-relative advantages are computed at the same writing position. For completion token $y_{g,t,k}$, the clipped GRPO objective is
\begin{equation}
\mathcal{J}(\theta)=\mathbb{E}_{g,t,k}\!\left[
\min\!\left(\rho_{g,t,k}\hat A_{g,t},
\operatorname{clip}(\rho_{g,t,k},1-\varepsilon,1+\varepsilon)\hat A_{g,t}\right)
-\beta d^{\mathrm{KL}}_{g,t,k}\right],
\label{eq:sc-grpo}
\end{equation}
We first average over tokens within each completion and then over samples with nonzero advantages; $\varepsilon=0.20$, $\beta=0.10$, and we use
\begin{equation}
d_{g,t,k}^{\mathrm{KL}}
=\exp(\delta_{g,t,k})-\delta_{g,t,k}-1,\qquad
\delta_{g,t,k}=\log\pi_{\mathrm{ref}}-\log\pi_\theta,
\end{equation}
$\pi_{\mathrm{ref}}$ is the frozen SFT policy. If the mean KL exceeds 0.02, the update is skipped.

\paragraph{Training data and hyperparameters.}
The Writer training set contains 15 multi-party networks and 73 writing segments (a writing segment is a Writer input unit, not the number of actions output by the model). To prevent sessions or windows from the same network appearing in both training and evaluation, train and held-out sets are separated at the complete-network level. We select the 10 shortest real SocialMemBench networks by message count, containing 50--101 messages, five sessions per network, and 5--12 terminal QA questions; they do not overlap with the 10 held-out SocialMemBench evaluation networks. We additionally use three manually authored UPDATE networks (30, 35, and 40 messages, covering state progression, UPDATE chains, source/owner separation, PERSON/GROUP scope, and temporal changes) and two manually authored NOOP networks (2--3 greeting or confirmation messages with no persistent fact).

Table~\ref{tab:r1-data-composition} gives a reproducible inventory of the training networks. Although the training inventory contains only 15 complete networks, each network is processed through multiple Writer segments, yielding 73 writing segments, 89 terminal QA items, and 452 supervised actions (430 ADD, 20 UPDATE, and 2 NOOP). The effective training signal therefore comes from repeated Writer decisions within each network rather than from 15 isolated samples. Writer trajectories for real networks are generated by DeepSeek-V4-Flash under the fixed Writer contract; manual networks are constructed directly by the experiment script with the same action and state contract and are not synthesized by a model. Each entry is isolated at the complete-network level, so messages, transitions, and terminal QA questions from one network never cross train and held-out sets.

\begin{table}[!t]
\caption{Composition of the R1 Writer training data. Real networks come from the SocialMemBench training side, while manual networks cover UPDATE/NOOP boundaries. A ``writing segment'' is a Writer input unit partitioned by session, not a count of ADD/UPDATE/NOOP actions; the QA column gives the available terminal questions for each network.}
\label{tab:r1-data-composition}
\centering
\scriptsize
\setlength{\tabcolsep}{3pt}
\begin{tabular}{llrrr}
\toprule
\textbf{Source} & \textbf{network ID} & \textbf{Messages} & \textbf{Writing segments} & \textbf{QA} \\
\midrule
Real & \texttt{grp\_0d1e2f3a} & 50 & 5 & 6 \\
Real & \texttt{grp\_5e6f7a8b} & 50 & 5 & 8 \\
Real & \texttt{grp\_6f7a8b9c} & 50 & 5 & 7 \\
Real & \texttt{grp\_7a8b9c0d} & 50 & 5 & 5 \\
Real & \texttt{grp\_8b9c0d1e} & 50 & 5 & 6 \\
Real & \texttt{grp\_9c0d1e2f} & 50 & 5 & 6 \\
Real & \texttt{grp\_4d5e6f7a} & 53 & 5 & 9 \\
Real & \texttt{grp\_3c4d5e6f} & 61 & 5 & 8 \\
Real & \texttt{grp\_2b3c4d5e} & 67 & 5 & 10 \\
Real & \texttt{grp\_1a2b3c4d} & 101 & 5 & 12 \\
Manual UPDATE & \texttt{manual\_update\_alpha} & 30 & 6 & 4 \\
Manual UPDATE & \texttt{manual\_update\_beta} & 35 & 7 & 3 \\
Manual UPDATE & \texttt{manual\_update\_gamma} & 40 & 8 & 3 \\
Manual NOOP & \texttt{manual\_noop\_greeting} & 2 & 1 & 1 \\
Manual NOOP & \texttt{manual\_noop\_ack} & 3 & 1 & 1 \\
\midrule
\rowcolor{gray!10}
\textbf{Total} & 15 networks & -- & 73 & 89 \\
\bottomrule
\end{tabular}
\end{table}

The 10 real networks produce 420 ADD and 3 UPDATE actions; the five manual networks produce 10 ADD, 17 UPDATE, and 2 NOOP actions, for 430 ADD, 20 UPDATE, and 2 NOOP actions in total. These action counts span all message positions and should not be added to or exchanged with the 73 writing segments in the table. Both RL and SFT Writers use the locally deployable Qwen2.5-3B. Real networks use DeepSeek-V4-Flash under the current Writer contract to generate teacher trajectories; these trajectories are proxy supervision rather than manually annotated item-level gold labels. Each network is sampled twice with group size $G=8$; at most four terminal QA questions are fixed for each training network. After SFT, we start from the epoch-10 checkpoint and run 30 rollout rounds. Each policy update uses two PPO epochs, learning rate $10^{-6}$, gradient-norm cap 1.0, temperature 0.80, top-$p=0.95$, clip 0.20, KL coefficient 0.10, and maximum generation length 2048. Terminal QA during training is used only for reward computation; the main benchmark evaluation still uses the fixed answerer and scoring protocol specified in the paper.

\paragraph{Controlled results.}
Under the same no-ASK query configuration as the main experiments (System~1 recall-$n=40$ with final top-$k=10$, System~2 $k=2$/source-$k=1$) and with the query and answer modules frozen, the correct counts for the three random seeds are 174/305, 175/305, and 176/305 for SFT (epoch 10), and 206/305, 208/305, and 210/305 for the RL Writer (30 steps). The corresponding mean accuracies are $57.38\pm0.33\%$ and $68.20\pm0.66\%$ (sample standard deviation), an average gain of 10.82 percentage points, or 33 additional correct answers. Under the same protocol, the three LLM writer runs with a DeepSeek-V4-Flash Writer obtain 216/305, 218/305, and 220/305, for a mean accuracy of $71.48\pm0.66\%$; the RL Writer is 3.28 points behind and reaches 95.4\% of the LLM writer accuracy. This comparison shows that the small Writer approaches the large-model reference, without being interpreted as evidence of cross-domain RL generalization.

\section{Complete Category Results on the Three Multi-Party Benchmarks}
\label{app:multibench-detailed}

This section gives the category results underlying the main summary tables; the main experiments contain 745, 1,031, and 2,400 questions, respectively. In all tables, $^\dagger$ denotes ASK disabled.

\subsection{Five Analysis Dimensions: Definitions and Typical Errors}
\label{app:five-challenges}
Here, the five dimensions form an analysis framework derived from question requirements and recurring error patterns; they are not independently human-annotated diagnostic labels. They organize related failure modes across the three benchmarks without defining a new validated taxonomy.

The five dimensions are not mutually exclusive question labels and do not alter the original annotations of the benchmarks. We retain each benchmark's official categories as quantitative reporting units, then map the dominant failure modes in each question to one or more diagnostic dimensions. The dimensions therefore explain why a question is missed rather than define a new aggregate score. The mapping uses the requirements in the question, object/scope/time constraints in the reference answer, and typical error cases, without additional manual gold labels beyond the test-set answers.

\paragraph{Full-member coverage.} Questions asking for "every person", "all participants", or a group member list require evidence for every roster member or an explicit statement that information is missing. A typical error is that global top-$k$ retrieval returns only frequent members and misses infrequent or late-arriving members.

\paragraph{Information attribution.} Questions require distinguishing the message source (who provides or expresses it) from the owner (whom the content concerns), while preserving the direction of reports, evaluations, and stances. A typical error is rewriting Alice's judgment about Bob as Bob's self-report.

\paragraph{Personal/group scope.} Questions require distinguishing an individual preference or exception, a two-person relation, a multi-person event, and a group norm or consensus. Typical errors elevate a minority opinion to a group conclusion or use a group rule to overwrite an individual exception.

\paragraph{Term and event disambiguation.} Questions require locating the correct branch among same-named terms, similar topics, or parallel events using participants, channels, time, and verbatim clues. A typical error is merging different projects, meetings, or decisions into one event.

\paragraph{Temporal updates and multi-hop reasoning.} Questions require selecting the requested point along a state-change chain and composing multiple pieces of evidence across members and events. Typical errors use an obsolete state as the current one or return only one fact from a chain without the connections needed for intermediate reasoning.

Table~\ref{tab:problem-mapping} maps the five problem dimensions to \method{} mechanisms; the main method section describes their implementation during writing and querying.
\begin{table}[!t]
\caption{Diagnostic dimensions and mechanisms in \method{}.}
\label{tab:problem-mapping}
\centering\small
\setlength{\tabcolsep}{4pt}
\begin{tabular}{>{\raggedright\arraybackslash}p{2.2cm}>{\raggedright\arraybackslash}p{5.1cm}>{\raggedright\arraybackslash}p{5.4cm}}
\toprule
\textbf{Challenge} & \textbf{Common gaps in summaries, event hierarchies, and graph memory} & \textbf{Corresponding mechanism in \method{}} \\
\midrule
Full-member coverage & Global top-$k$ and high-frequency nodes can retrieve much relevant evidence but do not guarantee one row per person & Deterministic roster, row-first System~2 retrieval, and explicit empty rows \\
Information attribution & Untyped facts or entity edges can merge ``who said it'' with ``whom it concerns'' & The profile layer and separate source/owner fields \\
Personal/group scope & Group summaries can elevate minority opinions to consensus, while person nodes do not express group norms well & PERSON core/profile and GROUP interaction/insight \\
Term and event disambiguation & Similar topics or graph connectivity can incorrectly merge same-named issues and parallel events & Fix person/GROUP rows first, then disambiguate within rows using events, relations, and provenance \\
Temporal updates and multi-hop reasoning & Overwrite updates lose history, while retaining every node may return stale states & Non-destructive update chains, head/full modes, and query-time evidence composition \\
\bottomrule
\end{tabular}
\end{table}

\begin{table}[!t]
\caption{Mapping between the five diagnostic dimensions and official categories of the three benchmarks. One official category may involve multiple dimensions, so this is not a mutually exclusive labeling scheme.}
\label{tab:challenge-taxonomy-mapping}
\centering
\scriptsize
\setlength{\tabcolsep}{3pt}
\begin{tabular}{>{\raggedright\arraybackslash}p{2.55cm}>{\raggedright\arraybackslash}p{6.1cm}>{\raggedright\arraybackslash}p{4.2cm}}
\toprule
\textbf{Diagnostic dimension} & \textbf{Typical official categories/question types} & \textbf{Primary diagnostic cues} \\
\midrule
Full-member coverage & SocialMem Q3/Q9; GroupMem per-member, outlier-member, and full-enumeration questions; EverMem multi-role coverage questions & Whether every target roster member is covered and infrequent, late-arriving, or departed members are handled correctly \\
Information attribution & SocialMem Q3/Q4/Q5; GroupMem user-implicit and relation questions; EverMem Role/Style questions & Whether source, owner, described person, audience, and observer remain consistent \\
Personal/group scope & SocialMem Q2/Q6/Q7; GroupMem group-dynamics questions; EverMem Const/Proact questions & Whether personal preferences, two-person relations, multi-person events, and group norms are distinguished \\
Term and event disambiguation & GroupMem term-ambiguity; SocialMem similar-issue/parallel-event questions; EverMem Multi/Role questions & Whether participant, channel, time, and event clues point to the same branch \\
Temporal updates and multi-hop reasoning & GroupMem temporal/knowledge-update/multi-hop; SocialMem Q7/Q8; EverMem Temp/Update/Multi & Whether the requested time is selected and evidence chains across members or events are connected \\
\bottomrule
\end{tabular}
\end{table}

\subsection{Category Definitions for the Three Benchmarks}
\label{app:benchmark-categories}
\paragraph{GroupMemBench categories.} GroupMemBench contains 745 questions in six categories: abstention (recognizing insufficient information and avoiding speculation), user-implicit (inferring behavior or preferences not directly stated by the user), multi-hop (composing answers across multiple memories), temporal (locating or comparing states by time), knowledge-update (handling knowledge updates and obsolete information), and term-ambiguity (disambiguating terms or expressions)~\citep{yang2026groupmembench}.

\paragraph{SocialMemBench categories.} SocialMemBench contains 1,031 questions: Q1 single-person recall and behavior patterns; Q2 group decisions and shared expectations; Q3 per-person stances; Q4 speaker attribution; Q5 cross-person beliefs; Q6 group norms and individual exceptions; Q7 two-person relations and shared history; Q8 temporal evolution; and Q9 outlier and departed-member information~\citep{owolabi2026socialmembench}.

\paragraph{EverMemBench categories.} The main EverMemBench analysis uses the nine memory behaviors published by EverMind-AI:\textit{Single}, \textit{Multi}, \textit{Temp}, \textit{Const}, \textit{Proact}, \textit{Update}, \textit{Style}, \textit{Skill}, and \textit{Role}~\citep{hu2026evermembench,evermind2026multiround}. Main-text Table~\ref{tab:evermem-cross-config} reports the nine category accuracies and the question-count-weighted overall accuracy over 2,400 questions; multiple-choice/open-ended format is no longer used as the primary classification.

\subsection{Binary Accuracy with DeepSeek-V4-Flash as the Answer Model}
The following tables report the DeepSeek-V4-Flash answerer configuration, while question-level binary results are judged uniformly by GPT-4o-mini; DeepSeek-V4-Flash is not the judge. Tables~\ref{tab:gm-acc-deepseek} and~\ref{tab:sm-acc-deepseek} give the category results for GroupMemBench and SocialMemBench, respectively. The main nine-behavior EverMemBench analysis is in main-text Table~\ref{tab:evermem-cross-config}.

\paragraph{GroupMemBench configuration.} GroupMemBench has 745 questions in the six categories defined above. DeepSeek-V4-Flash generates answers and GPT-4o-mini provides the question-level binary judge. Each system follows its official code, recommended configuration, and official prompt; we standardize only the data, metric definition, judge, and evaluation interface, with speaker/source fields retained where supported.
\begin{table}[!t]
\caption{GroupMemBench category accuracy (\%).}
\label{tab:gm-acc-deepseek}
\centering\scriptsize
\resizebox{\textwidth}{!}{%
\begin{tabular}{lrrrrrrr}
\toprule
\textbf{Method} & \textbf{Abst.} & \textbf{Implicit} & \textbf{Multi-hop} & \textbf{Temporal} & \textbf{K.-update} & \textbf{Term-amb.} & \textbf{Total} \\
\midrule
BM25 & 89.9 & 46.9 & 41.8 & 41.4 & 23.4 & 15.1 & 44.6 \\
Embed & 87.8 & 40.8 & 25.8 & 17.3 & 20.6 & 17.0 & 34.5 \\
Mem0 & 88.5 & 10.2 & 7.1 & 2.5 & 5.6 & 9.4 & 21.6 \\
A-MEM & 80.6 & 22.4 & 15.9 & 5.6 & 13.1 & 25.5 & 27.1 \\
HippoRAG & 77.7 & 22.4 & 18.7 & 5.6 & 11.2 & 25.5 & 27.0 \\
\rowcolor{gray!10}
\method{}$^\dagger$ & 71.2 & 44.9 & 40.7 & 49.4 & 33.6 & 36.8 & \textbf{47.0} \\
\rowcolor{gray!10}
\method{} & 73.4 & 51.0 & 43.4 & 50.6 & 33.6 & 31.1 & \textbf{47.9} \\
\bottomrule
\end{tabular}%
}
\end{table}

\paragraph{SocialMemBench configuration.} SocialMemBench has 1,031 questions partitioned into Q1--Q9. DeepSeek-V4-Flash generates answers and GPT-4o-mini supplies every binary verdict under the standardized data, metric, judge, and evaluation-interface protocol. Full context is feasible only for SocialMemBench; the other benchmark contexts are too long for full-context input.
\begin{table}[!t]
\caption{SocialMemBench Q1--Q9 category accuracy (\%).}
\label{tab:sm-acc-deepseek}
\centering\scriptsize
\resizebox{\textwidth}{!}{%
\begin{tabular}{lrrrrrrrrrr}
\toprule
\textbf{Method} & \textbf{Q1} & \textbf{Q2} & \textbf{Q3} & \textbf{Q4} & \textbf{Q5} & \textbf{Q6} & \textbf{Q7} & \textbf{Q8} & \textbf{Q9} & \textbf{Total} \\
\midrule
BM25 & 25.3 & 24.7 & 4.5 & 66.2 & 38.0 & 25.9 & 29.8 & 20.6 & 10.0 & 28.6 \\
Embed & 39.4 & 28.4 & 13.6 & 62.5 & 53.3 & 46.3 & 40.9 & 25.6 & 40.0 & 38.1 \\
Mem0 & 20.9 & 14.8 & 4.5 & 11.3 & 3.3 & 25.9 & 9.4 & 12.2 & 10.0 & 13.7 \\
A-MEM & 60.6 & 76.5 & 9.1 & 80.0 & 65.2 & 51.8 & 53.0 & 45.0 & 50.0 & 56.8 \\
HippoRAG & 60.6 & 71.6 & 4.5 & 68.8 & 63.0 & 48.1 & 53.0 & 47.7 & 60.0 & 55.9 \\
Full context & 74.3 & 51.9 & 18.2 & 73.8 & 77.2 & 53.7 & 77.3 & 68.3 & 70.0 & 69.4 \\
\rowcolor{gray!10}
\method{}$^\dagger$ & 73.9 & 69.1 & 9.1 & 90.0 & 78.3 & 46.3 & 76.2 & 59.9 & 70.0 & \textbf{69.2} \\
\rowcolor{gray!10}
\method{} & 69.1 & 70.4 & 4.5 & 80.0 & 79.3 & 53.7 & 65.7 & 56.1 & 70.0 & 64.9 \\
\bottomrule
\end{tabular}%
}
\end{table}

\subsection{Category-Level token-F1}

token-F1 is not equivalent to question-level binary accuracy. The results below are category means of question-level token-F1; \method{} F1 is computed from outputs using the same token-matching rule as the main experiments. Tables~\ref{tab:gm-f1} and~\ref{tab:sm-f1} give category-level token-F1 for the two multi-party benchmarks; EverMemBench token-F1 is reported only as an overall supplementary metric in the main table.

\paragraph{GroupMemBench token-F1.} The following table gives token-F1 on the same six categories. The rows follow their official code, recommended configuration, and official prompt within the DeepSeek-V4-Flash answerer configuration; we standardize the data, token-F1 definition, and evaluation interface. token-F1 is computed directly against references and does not use the GPT-4o-mini judge.

\begin{table}[!t]
\caption{GroupMemBench category token-F1 (\%).}
\label{tab:gm-f1}
\centering\scriptsize
\resizebox{\textwidth}{!}{%
\begin{tabular}{lrrrrrrr}
\toprule
\textbf{Method} & \textbf{Abst.} & \textbf{Implicit} & \textbf{Multi-hop} & \textbf{Temporal} & \textbf{K.-update} & \textbf{Term-amb.} & \textbf{Total} \\
\midrule
BM25 & 25.6 & 33.5 & 25.8 & 36.4 & 24.2 & 2.0 & 25.0 \\
Embed & 24.2 & 27.0 & 17.1 & 14.2 & 22.4 & 4.4 & 17.4 \\
Mem0 & 26.0 & 7.0 & 5.0 & 0.0 & 13.0 & 1.0 & 9.0 \\
A-MEM & 15.0 & 15.8 & 8.3 & 5.9 & 19.7 & 4.8 & 10.6 \\
HippoRAG & 14.6 & 12.4 & 11.3 & 6.2 & 19.7 & 4.5 & 11.1 \\
\rowcolor{gray!10}
\method{}$^\dagger$ & 14.5 & 26.6 & 22.3 & 49.7 & 26.5 & 5.1 & 25.2 \\
\rowcolor{gray!10}
\method{} & 15.1 & 25.6 & 26.3 & 50.9 & 26.7 & 4.4 & \textbf{26.5} \\
\bottomrule
\end{tabular}%
}
\end{table}

\paragraph{SocialMemBench token-F1.} token-F1 is reported by Q1--Q9 below. The answerer is DeepSeek-V4-Flash; each row follows its official code, recommended configuration, and official prompt, while the data, token-F1 definition, and evaluation interface are standardized. Values are computed directly against references rather than derived from GPT-4o-mini judgments. Full context appears only for this benchmark.

\begin{table}[!t]
\caption{SocialMemBench Q1--Q9 category token-F1 (\%).}
\label{tab:sm-f1}
\centering\scriptsize
\resizebox{\textwidth}{!}{%
\begin{tabular}{lrrrrrrrrrr}
\toprule
\textbf{Method} & \textbf{Q1} & \textbf{Q2} & \textbf{Q3} & \textbf{Q4} & \textbf{Q5} & \textbf{Q6} & \textbf{Q7} & \textbf{Q8} & \textbf{Q9} & \textbf{Total} \\
\midrule
BM25 & 13.8 & 10.3 & 12.1 & 38.1 & 18.7 & 14.5 & 15.7 & 12.2 & 8.2 & 15.7 \\
Embed & 15.2 & 12.5 & 10.0 & 39.8 & 20.8 & 18.4 & 18.6 & 15.8 & 7.8 & 18.1 \\
Mem0 & 12.0 & 7.0 & 16.0 & 7.0 & 10.0 & 12.0 & 11.0 & 10.0 & 10.0 & 11.0 \\
A-MEM & 22.3 & 62.4 & 24.1 & 67.3 & 29.2 & 51.9 & 23.4 & 24.7 & 20.9 & 31.9 \\
HippoRAG & 21.2 & 60.1 & 23.4 & 55.7 & 30.6 & 49.0 & 23.1 & 25.0 & 19.0 & 30.6 \\
Full context & 22.6 & 26.8 & 8.5 & 47.3 & 30.0 & 31.9 & 24.1 & 23.3 & 16.9 & 26.1 \\
\rowcolor{gray!10}
\method{}$^\dagger$ & 27.9 & 5.9 & 26.5 & 35.4 & 34.7 & 4.6 & 30.8 & 30.9 & 33.9 & 27.4 \\
\rowcolor{gray!10}
\method{} & 22.8 & 56.5 & 25.0 & 70.9 & 30.9 & 52.4 & 24.9 & 26.0 & 22.4 & \textbf{32.7} \\
\bottomrule
\end{tabular}%
}
\end{table}

\paragraph{EverMemBench token-F1.} EverMemBench token-F1 is reported only as an overall supplementary metric. The main table reports question-level binary accuracy by nine behaviors; token-F1 is not conflated with category accuracy or question format.

\subsection{GPT-5.6-luna Model Configuration}

This configuration uses GPT-5.6-luna across memory construction, retrieval, and answer generation, while retaining GPT-4o-mini as the question-level judge, to assess cross-model robustness. BM25 and Embed use single-message top-10 retrieval, and Embed uses the local \texttt{all-MiniLM-L6-v2}. Full context is feasible only on SocialMemBench. Overall results are in main-text Table~\ref{tab:main-results}; category details for GroupMemBench and SocialMemBench are given in Tables~\ref{tab:gm-acc-luna} and~\ref{tab:sm-acc-luna}, respectively. Only the overall EverMemBench result is reported under this model configuration.

\paragraph{GroupMemBench category results.} This table uses the six categories defined above and contains 745 questions. The complete pipeline uses GPT-5.6-luna for memory construction, retrieval-time evidence interpretation, and answer generation, while GPT-4o-mini supplies the question-level binary judgments. Each method follows its official code, recommended configuration, and official prompt; we standardize only the data, metric definition, judge, and evaluation interface. BM25 and Embed use single-message top-10 evidence; SpeakerMem uses System~1 top-10 and System~2 owner-2 plus source-1. The entries are category-wise binary accuracies.

\begin{table}[!t]
\caption{GroupMemBench category accuracy with GPT-5.6-luna (\%).}
\label{tab:gm-acc-luna}
\centering\scriptsize
\resizebox{\textwidth}{!}{%
\begin{tabular}{lrrrrrrr}
\toprule
\textbf{Method} & \textbf{Abst.} & \textbf{Implicit} & \textbf{Multi-hop} & \textbf{Temporal} & \textbf{K.-update} & \textbf{Term-amb.} & \textbf{Total} \\
\midrule
\rowcolor{gray!10}
\method{} & 57.6 & 49.0 & 42.3 & 42.0 & 26.2 & 38.7 & 42.7 \\
BM25 & 71.9 & 49.0 & 43.4 & 56.8 & 27.1 & 18.9 & 46.2 \\
Embed & 70.5 & 44.9 & 29.7 & 25.3 & 19.6 & 25.5 & 35.3 \\
Mem0 & 82.0 & 20.4 & 6.6 & 0.6 & 0.9 & 12.3 & 20.27 \\
A-MEM & 65.5 & 30.6 & 21.4 & 1.2 & 13.1 & 34.9 & 26.58 \\
HippoRAG & 64.7 & 28.6 & 22.5 & 4.9 & 13.1 & 33.0 & 27.11 \\
\bottomrule
\end{tabular}%
}
\end{table}

\paragraph{SocialMemBench category results.} This table uses the Q1--Q9 partition defined above and contains 1,031 questions. The answerer is GPT-5.6-luna and GPT-4o-mini is the binary judge under the standardized data, metric, judge, and evaluation-interface protocol. Full context is shown only here because the longer benchmark inputs do not fit a practical context. The entries are category-wise binary accuracies.

\begin{table}[!t]
\caption{SocialMemBench category accuracy with GPT-5.6-luna (\%).}
\label{tab:sm-acc-luna}
\centering\scriptsize
\resizebox{\textwidth}{!}{%
\begin{tabular}{lrrrrrrrrrr}
\toprule
\textbf{Method} & \textbf{Q1} & \textbf{Q2} & \textbf{Q3} & \textbf{Q4} & \textbf{Q5} & \textbf{Q6} & \textbf{Q7} & \textbf{Q8} & \textbf{Q9} & \textbf{Total} \\
\midrule
\rowcolor{gray!10}
\method{} & 68.7 & 72.8 & 4.5 & 83.8 & 80.4 & 53.7 & 58.6 & 56.9 & 80.0 & 64.4 \\
BM25 & 25.7 & 33.3 & 4.5 & 75.0 & 35.9 & 37.0 & 30.9 & 16.8 & 40.0 & 30.0 \\
Embed & 35.7 & 39.5 & 4.5 & 66.2 & 52.2 & 53.7 & 34.8 & 21.8 & 30.0 & 36.4 \\
Mem0 & 20.1 & 21.0 & 9.1 & 17.5 & 0.0 & 31.5 & 7.2 & 11.5 & 10.0 & 13.97 \\
A-MEM & 54.6 & 63.0 & 4.5 & 70.0 & 54.3 & 38.9 & 43.1 & 32.1 & 30.0 & 46.56 \\
HippoRAG & 61.4 & 77.8 & 4.5 & 77.5 & 69.6 & 61.1 & 62.4 & 51.9 & 70.0 & 61.30 \\
Full context & 71.1 & 79.0 & 4.5 & 82.5 & 71.7 & 66.7 & 72.9 & 69.8 & 100.0 & 71.3 \\
\bottomrule
\end{tabular}%
}
\end{table}

\paragraph{EverMemBench category results.} This model configuration is used for overall cross-model robustness analysis; the primary report of the nine behaviors uniformly uses the gpt-4.1-mini/Gemini-3-Flash configuration in main-text Table~\ref{tab:evermem-cross-config}.

\subsection{Complete Path and Hierarchy Ablations}
\label{app:complete-ablations}
\begin{table}[!t]
\caption{Accuracy (\%) for dual-track path and hierarchy ablations. SM, GM, and EM denote SocialMemBench, GroupMemBench, and EverMemBench. All configurations use the same System~1 top-$k=10$ and System~2 $k=2$/source-$k=1$. Row names match Figure~4: w/o S2 removes all four System~2 derived layers, w/o S1 removes the verbatim track, w/o S2-group removes Interaction and Insight, w/o S2-person removes Core and Profile, and SpeakerMem denotes the full dual-track system. The last three $\Delta$ columns are relative to the best full result for each benchmark.}
\label{tab:ablations}
\centering
\small
\renewcommand{\arraystretch}{1.04}
\setlength{\tabcolsep}{3pt}
\begin{tabularx}{\textwidth}{@{}l>{\centering\arraybackslash}X>{\centering\arraybackslash}X>{\centering\arraybackslash}X>{\centering\arraybackslash}X>{\centering\arraybackslash}X>{\centering\arraybackslash}X@{}}
\toprule
\textbf{Configuration} & \textbf{SM} & \textbf{GM} & \textbf{EM} & \multicolumn{3}{c}{\textbf{$\Delta$ vs. best full}} \\
 & \textbf{Acc.} & \textbf{Acc.} & \textbf{Acc.} & \textbf{SM} & \textbf{GM} & \textbf{EM} \\
\midrule
w/o S2 & 54.80 & 42.01 & 48.92 & $-14.40$ & $-5.89$ & $-12.98$ \\
w/o S1 & 42.58 & 33.15 & 38.33 & $-26.62$ & $-14.75$ & $-23.57$ \\
w/o S2-group & 65.37 & 43.22 & 55.92 & $-3.83$ & $-4.68$ & $-5.98$ \\
w/o S2-person & 63.72 & 41.07 & 54.88 & $-5.48$ & $-6.83$ & $-7.02$ \\
\midrule
SpeakerMem$^*$ & 69.2 & 47.0 & 60.5 & $0.0$ & $-0.9$ & $-1.4$ \\
\rowcolor{gray!10}
SpeakerMem & \textbf{64.9} & \textbf{47.9} & \textbf{61.9} & $-4.3$ & $0.0$ & $0.0$ \\
\bottomrule
\end{tabularx}
\smallskip
\noindent\parbox{\linewidth}{\footnotesize\textit{Supplementary retrieval ablation:} SpeakerMem$^*$ across the three domains is 69.2/47.0/60.5; SpeakerMem is 64.9/47.9/61.9; $^*$ means ASK disabled; SpeakerMem denotes the default ASK-enabled configuration.}
\end{table}

\section{Full LoCoMo Category Breakdown}
\label{app:locomo-detailed}

LoCoMo contains 1,986 questions in five categories~\citep{maharana2024locomo}. Table~\ref{tab:locomo-full-categories} summarizes \method{} results by category. The categories are Single-hop (one-step factual recall), Adversarial/AD (adversarial or misleading questions), Temporal (temporal relations and state changes), Multi-hop (reasoning over multiple dialogue facts), and Open-domain (requiring open-domain knowledge or cross-context reasoning).

\begin{table}[!t]
\caption{Question-level binary accuracy of \method{} on all five LoCoMo categories.}
\label{tab:locomo-full-categories}
\centering\small
\begin{tabular}{lrr}
\toprule
\textbf{Category} & \textbf{Correct/total} & \textbf{Accuracy (\%)} \\
\midrule
Single-hop & 655/841 & 77.88 \\
Adversarial/AD & 370/446 & 82.96 \\
Temporal & 227/321 & 70.72 \\
Multi-hop & 116/282 & 41.13 \\
Open-domain & 39/96 & 40.62 \\
\midrule
\rowcolor{gray!10}
All questions & 1407/1986 & \textbf{70.85} \\
Excluding AD & 1037/1540 & 67.34 \\
\bottomrule
\end{tabular}
\end{table}

The Single-hop and Temporal results show that this structure remains effective for two-person long-term conversations. The sharp drop on Multi-hop and Open-domain reveals that cross-segment composition and external knowledge remain major weaknesses. The main-text Table~\ref{tab:locomo-compatibility} gives the corresponding category-level positions of published baselines.

\FloatBarrier
\section{\texorpdfstring{Retrieval Top-$k$ Sensitivity}{Retrieval Top-k Sensitivity}}
\label{app:topk-sensitivity}

\paragraph{Purpose and controlled setting.}
This experiment examines how the two retrieval budgets jointly affect answer quality, distinguishing additional verbatim coverage from additional derived records. We reuse the fixed memory store constructed for the main experiments on SocialMemBench held-out eval10 (10 networks, 305 questions), using dual-track querying without ASK (for this sensitivity study, S1 recall-$n$ is set equal to S1 top-$k$, so no larger candidate pool is recalled before Select). We evaluate all nine combinations of S1 top-$k\in\{5,10,20\}$ and S2 $k\in\{2,4,6\}$, with S2 \texttt{source\_k}=1 throughout. The two varied budgets concern S1 raw candidates and S2 owner/row retrieval, respectively; the source budget does not vary. No RL/SFT/Qwen Writer is used, so this is a retrieval-budget study with fixed memory. Acc is binary accuracy evaluated by GPT-4o-mini, and token-F1 is computed from per-question predictions and references. Both metrics are reported as percentages.
\begin{figure}[!t]
  \centering
  \includegraphics[width=\linewidth]{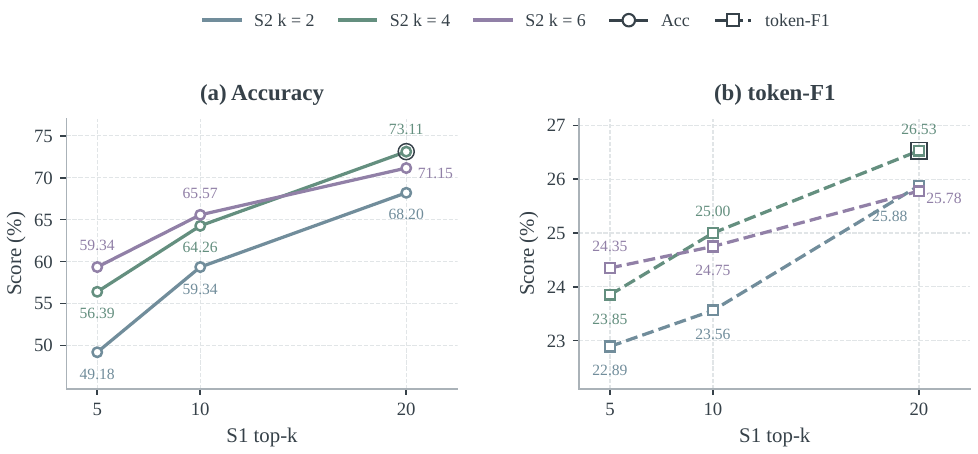}
  \caption{Retrieval-budget response curves. Accuracy (left) and token-F1 (right) are plotted against S1 top-$k$; colors distinguish S2 $k=2,4,6$, and solid circles versus dashed squares identify the metrics. Panels use separate y-axis ranges to reveal smaller token-F1 changes; visual slopes should therefore not be compared across metrics. Outlined markers identify the best observed configuration for each metric. Lines connect measured configurations only. The setting uses the main-experiment memory store, 305 questions, no ASK, and S2 source-$k=1$.}
  \label{fig:topk-sensitivity-curves}
\end{figure}

\paragraph{S1: additional coverage helps, but returns per budget slot diminish.}
S1 preserves original messages with speaker, time, and local context, supplying verifiable evidence for S2 records and wording that compression may omit. Increasing S1 top-$k$ from 5 to 10 and then to 20 improves both Acc and token-F1 at every tested S2 budget, indicating that small verbatim budgets still miss useful information. At S2 $k=4$, for example, Acc rises from 56.39\% to 64.26\% and 73.11\%, while token-F1 rises from 23.85\% to 25.00\% and 26.53\%. However, the first increment adds five candidate slots and the second adds ten, so their total gains alone do not measure diminishing returns. For S2 $k=2,4,6$, the average Acc gain per added S1 slot falls from 2.03, 1.57, and 1.25 percentage points over 5--10 to 0.89, 0.89, and 0.56 over 10--20. Thus, \emph{returns per S1 budget slot slow markedly, suggesting a trend toward saturation}: after high-value messages have been covered, additional candidates may increasingly repeat existing evidence or contribute weaker clues. Total Acc gains over 10--20 remain 5.58--8.86 points, while token-F1 marginal gains are smaller and less uniform.

\paragraph{S2: a moderate increase helps, whereas excess records can impair judgment.}
S2 supplies addressable derived states constrained by PERSON/GROUP scope, source/owner, event, and time. Increasing S2 $k$ from 2 to 4 improves both metrics at all three S1 budgets: Acc gains are 7.21, 4.92, and 4.91 points, and token-F1 gains are 0.96, 1.44, and 0.65 points. This is consistent with a moderate expansion recovering complementary participant, relation, and state clues. Increasing S2 further from 4 to 6 is less reliable. At S1 top-$k=10$, Acc rises only from 64.26\% to 65.57\%, while token-F1 falls from 25.00\% to 24.75\%. At S1 top-$k=20$, Acc falls from 73.11\% to 71.15\% and token-F1 from 26.53\% to 25.78\%, decreases of 1.96 and 0.75 points, respectively. One explanation consistent with the dual-track design is that excess derived records introduce duplicate facts, overlapping descriptions, or weaker relational and historical clues, diluting decisive evidence and increasing the answerer's judgment burden. S2 should therefore retain a moderate set of complementary records rather than maximize quantity. The decline is directly observed; redundant or distracting records are one plausible mechanism.

\begin{figure}[!t]
  \centering
  \includegraphics[width=0.88\linewidth]{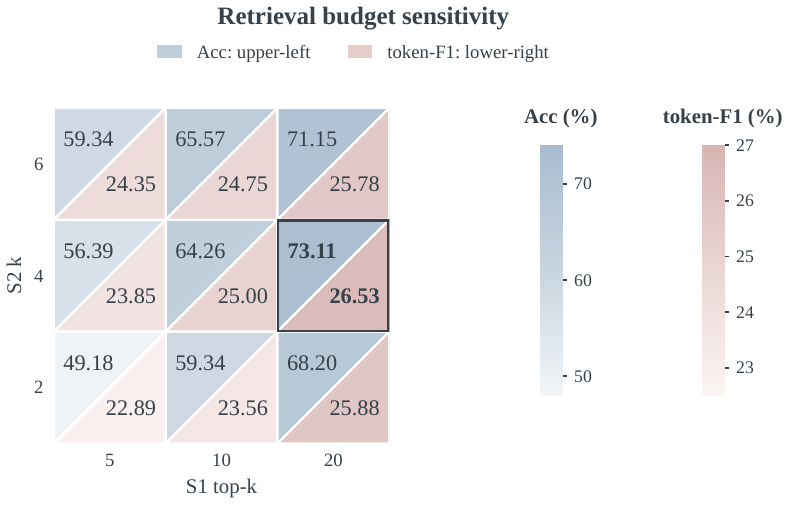}
  \caption{Joint S1/S2 configuration grid. The horizontal axis enumerates discrete S1 top-$k$ settings and the vertical axis gives S2 $k$. Each cell reports exact percentages: the upper-left blue triangle encodes Acc and the lower-right rose triangle encodes token-F1. Independent color scales prevent direct cross-metric comparisons of shade. The outlined cell, $(20,4)$, is the best observed configuration for both metrics. These are the same nine observations as Figure~\ref{fig:topk-sensitivity-curves}, not a separate experiment.}
  \label{fig:topk-sensitivity-grid}
\end{figure}

\paragraph{Dual-track interaction: S2's benefit depends on verbatim coverage.}
Figure~\ref{fig:topk-sensitivity-grid} shows that the effect of extra S2 budget depends on S1. With S1 top-$k=5$, increasing S2 from 4 to 6 still improves Acc/token-F1 from 56.39/23.85 to 59.34/24.35, consistent with additional derived clues filling gaps in limited raw coverage. The same S2 increment decreases both metrics when S1 top-$k=20$: once S1 supplies more complete original evidence, the complementary value of additional structured records may no longer outweigh redundancy and distraction. This pattern supports separate budget control for the two tracks. S1 expands verifiable verbatim coverage, while S2 should provide a compact set of query-relevant states and relations. It also fits Anchor--Separate--Resolve--Compose: attribution, scope, and versions are constrained before evidence is composed, and the two budgets need not be maximized together.

\paragraph{Acc and token-F1 jointly inform budget selection.}
Acc assesses task-level correctness, whereas token-F1 measures surface token overlap, so the metrics need not move together. With S1 top-$k=10$, increasing S2 from 4 to 6 raises Acc but lowers token-F1. A single metric could therefore overlook changes in answer completeness, concision, or wording, although this discrepancy alone cannot identify the responsible factor. Among the nine configurations, S1 top-$k=20$ and S2 $k=4$ jointly maximize Acc (73.11\%) and token-F1 (26.53\%), suggesting a useful quality balance between sufficient verbatim evidence and a moderate structured budget. This is the best observed quality balance in the tested grid; the main experiments retain S1 top-$k=10$ and S2 $k=2$/source-$k=1$. Overall, the results favor monitoring marginal S1 returns and keeping S2 moderate.
\FloatBarrier

\section{Complete Prompts}
\label{app:prompts}

This section provides snapshots of the key prompts used in the experiments: Writer, Project, System~1 Select/Expand, Sufficiency/ASK, the frozen answerer, binary judging, and SocialMem MeanQ/MeanN scoring. Fields in braces are filled at runtime; except for the message wrapper required by the model API, the text below is neither summarized nor rewritten. This makes the module responsibilities and actual output formats in the main text directly auditable.

\begin{table}[!t]
\caption{Input/output formats and training boundaries of key prompts.}
\label{tab:prompt-inventory}
\centering
\scriptsize
\setlength{\tabcolsep}{1pt}
\begin{tabular}{>{\raggedright\arraybackslash}p{2.4cm}>{\raggedright\arraybackslash}p{4.1cm}>{\raggedright\arraybackslash}p{4.0cm}>{\raggedright\arraybackslash}p{3.1cm}}
\toprule
\textbf{Module} & \textbf{Input} & \textbf{Output format} & \textbf{Boundary} \\
\midrule
Writer & Roster, local message segment, and current derived-layer heads & \textsc{Add}/\textsc{Update}/\textsc{Noop} JSON; does not generate answers & Qwen2.5-3B; trained with SFT/R1 \\
Project & Question and channel roster & issue, PERSON/GROUP rows, \texttt{head/full}, and relation constraints & Frozen; does not read global answer candidates \\
Select/Expand & Question and System~1 verbatim candidates & Necessity ranking and at most two adjacent-window anchors & Frozen; applies only to System~1 \\
Sufficiency/ASK & Truncated System~1 evidence & \texttt{enough} and an optional search term & Frozen; at most one supplementary retrieval \\
Answer & Question, final evidence from both paths, empty rows, and temporal mode & Final natural-language answer & Frozen; does not modify memory \\
Binary judge & Question, gold answer, and system answer & Question-level 0/1 decision and structured rationale & Both DeepSeek-V4-Flash and GPT-5.6-luna rows use GPT-4o-mini for judging; the judge for EverMemBench cross-configuration results is given in Table~\ref{tab:evermem-cross-config} \\
SocialMem scoring protocol & Question, gold answer, system answer, and scoring protocol & Partial scores used by MeanQ/MeanN & Both DeepSeek-V4-Flash and GPT-5.6-luna rows use GPT-4o-mini for judging; the judge for EverMemBench cross-configuration results is given in Table~\ref{tab:evermem-cross-config} \\
\bottomrule
\end{tabular}
\end{table}

\subsection{Prompt Inventory}
\label{app:prompt-inventory}

Table~\ref{tab:prompt-inventory} lists the input, output, and frozen boundary for each prompt; the following subsections preserve the complete copyable text. Except for the Writer, query, answer, and judging prompts are fixed in the main experiments and ablations.

\subsection{Write: Structured Writing}

The Writer input contains the channel roster, the latest state of derived memory, and a new local message segment. The model generates only \textsc{Add}, \textsc{Update}, or \textsc{Noop}; time, session, verbatim \texttt{from\_ids}, and update-chain fields are added by the system after the action passes validation. The complete prompt follows.

\promptinput{Prompt 1: Writer}{appendix/prompt_texts/writer.txt}

\subsection{Project: Query Projection}

Project outputs only \texttt{issue}, \texttt{rows}, and \texttt{tense}. \texttt{rows} determine the PERSON/GROUP views and \texttt{tense} determines whether to use only current nodes or expand update chains; channel and source/owner constraints are enforced by the roster and within-row retrieval.

\promptinput{Prompt 2: Project}{appendix/prompt_texts/project.txt}

\subsection{Select and Expand: Verbatim-Track Selection}

This prompt operates only on System~1 verbatim candidates. It returns both a necessity ranking and at most two adjacent-window expansion anchors. Multi-person coverage is handled by System~2 and is therefore not a condition for triggering Expand.

\promptinput{Prompt 3: System~1 Select / Expand}{appendix/prompt_texts/select_expand.txt}

\subsection{Sufficiency and ASK: Supplementary Retrieval Decision}

Sufficiency independently checks the System~1 evidence that will be passed to the answerer and outputs only \texttt{enough} and one optional search term. If \texttt{enough=false}, the system uses \texttt{ask} to retrieve and rerank once more; this module does not generate the final answer.

\promptinput{Prompt 4: Sufficiency / ASK}{appendix/prompt_texts/sufficiency_ask.txt}

\subsection{Answer: Frozen answerer}

The answerer receives speaker/time-labeled verbatim evidence and a sparse structured view organized by PERSON/GROUP rows. The prompt explicitly forbids replacing an individual position with a group conclusion and specifies how to interpret \texttt{head/full} and empty rows.

\promptinput{Prompt 5: Answer}{appendix/prompt_texts/answer.txt}

\subsection{Evaluation Prompts}

The question-level binary accuracy in the main tables uses the following fixed judging prompt. The user message always contains the question, gold answer, and system answer.

\promptinput{Prompt 6: Binary Judge}{appendix/prompt_texts/binary_judge.txt}

SocialMem MeanQ/MeanN use the benchmark scoring protocol with partial credit: MeanQ averages item-level scores, while MeanN first averages within dialogue networks and then averages across networks. The complete scoring protocol follows.

\promptinput{Prompt 7: SocialMem scoring rubric}{appendix/prompt_texts/socialmem_rubric_judge.txt}

\end{document}